\documentclass[10pt,twocolumn,letterpaper]{article}

\usepackage[pagenumbers]{cvpr} 

\usepackage[dvipsnames]{xcolor}

\definecolor{cvprblue}{rgb}{0.21,0.49,0.74}
\usepackage[pagebackref,breaklinks,colorlinks,citecolor=cvprblue]{hyperref}
\usepackage{multirow}
\usepackage{makecell}
\def\paperID{7056} 
\def\confName{CVPR}
\def\confYear{2024}

\title{MLPHand: Real Time Multi-View 3D Hand Mesh Reconstruction via MLP Modeling}

\author{
Jian Yang$^{1}$\thanks{Equal contribution.}, Jiakun Li$^{2*}$, Guoming Li$^{3}$, Zhen Shen$^{1}$,Huai-Yu Wu$^{1}$, Zhaoxin Fan$^{4}$\thanks{Corresponding Author}, Heng Huang$^{5}$
\\
$^{1}$Institute of Automation, Chinese Academy of Science,
$^{2}$Southern University of Science and Technology,\\
$^{3}$MBZUAI,
$^{4}$Renmin University of China, 
$^{5}$University of Maryland, College Park\\
}

\begin{document}
\maketitle
\begin{abstract}
Multi-view hand mesh reconstruction is a critical task for applications in virtual reality and human-computer interaction, but it remains a formidable challenge. Although existing multi-view hand reconstruction methods achieve remarkable accuracy, they typically come with an intensive computational burden that hinders real-time inference. To this end, we propose MLPHand, a novel method designed for real-time multi-view single hand reconstruction. MLPHand consists of two primary modules: (1) a lightweight MLP-based Skeleton2Mesh model that efficiently recovers hand meshes from hand skeletons, and (2) a multi-view geometry feature fusion prediction module that enhances the Skeleton2Mesh model with detailed geometric information from multiple views. Experiments on three widely used datasets demonstrate that MLPHand can reduce computational complexity by 90\% while achieving comparable reconstruction accuracy to existing state-of-the-art baselines.
\end{abstract}

\section{Introduction}
\label{sec:intro}

Hand mesh reconstruction holds a paramount position within the domains of virtual reality \cite{buckingham2021hand,kumar2015mujoco}  and augmented reality \cite{reifinger2007static,radkowski2012interactive}. It serves as a foundational component, not only enhancing the immersion of user experiences within the gaming industry but also underpinning a multitude of applications. Given captured signals (e.g., images/videos/point clouds) as input, advanced hand reconstruction methodologies typically employ deep learning models to simultaneously predict the hand's shape and pose. The ultimate goal is to recover a detailed hand mesh.

With this objective, monocular hand reconstruction methods~\cite{boukhayma20193d,kong2022identity,zimmermann2019freihand,zhou2020monocular,chen2021camera,chen2022mobrecon,tse2022collaborative,lin2021mesh,chen2021i2uv,lin2021end} have made noteworthy progress over the years. 
However, it is well-known that recovering 3D structures from a single image is an ill-posed problem, due to the depth ambiguity and self-occlusion.
To overcome these limitations, reconstructing an object using multi-view images has emerged as a promising approach. Studies such as \cite{wu2020multi,gordon2022flex,li20213d,zhang2021direct,dong2021shape} have demonstrated enhanced accuracy in human pose and shape estimation tasks by employing this strategy.
However, these studies have primarily focused on the reconstruction of full-body poses and shapes, overlooking the specific challenge of single-hand reconstruction.
To address this gap, POEM~\cite{yang2023poem} is recently proposed for multi-view single hand reconstruction and achieves impressive accuracy.
However, its intricate cross-set feature interaction imposes significant computational demands, consequently limiting its inference speed. Nevertheless, the fast inference time is crucial in hand reconstruction tasks, where efficiency is equally as important as accuracy in evaluating the efficacy of reconstruction methods. 

Therefore, a pivotal question arises: \textbf{can we develop a real-time hand reconstruction method without sacrificing accuracy?}
The answer is undoubtedly "Yes".
Motivated by this intuition, we present MLPHand, the first real-time multi-view hand reconstruction method. 
It contains two key innovations for achieving real-time inference. Firstly, we propose a light-weight Skeleton2Mesh model inspired by recent advancements in MLP-based geometry modeling~\cite{karunratanakul2021skeleton,chen2023hand,corona2022lisa,mihajlovic2022coap,deng2020nasa}. Furthermore, this model adopts a pure-MLP architecture with a custom-designed Tri-Axis modeled Per-Bone reconstruction scheme, resulting in an exceedingly simple network that supports real-time forward propagation for hand mesh recovery. Secondly, we propose another MLP-based multi-view geometry feature fusion prediction module to enhance the performance of Skeleton2Mesh model. This module infuses multi-view hand-related vision features into the Skeleton2Mesh model, thus improving the prediction of hand details without compromising inference speed. With these two key designs, MLPHand can accurately reconstruct the human hand from multi-view images while maintaining real-time performance, demonstrating its potential as a practical, efficient solution for multi-view hand reconstruction. 

To substantiate the efficacy and efficiency of MLPHand, we conduct extensive experiments on three widely adopted datasets~\cite{chao2021dexycb,yang2022oakink,hampali2021ho}. The results reveal that MLPHand not only achieves real-time performance, operating at 71 FPS on a 3090 GPU, but also marks a considerable decrease in both parameter size (75\%) and computational cost (90\%) compared to state-of-the-art methods.  Simultaneously, MLPHand upholds a level of performance accuracy that is comparable with existing baselines.
The main contributions of our work can be summarized as follows:

\begin{itemize}
  \item  We introduce, to the best of our knowledge, the first real-time multi-view human hand reconstruction method, termed MLPHand. This method operates at a speed of 71 FPS on a 3090 GPU, enabled by a 75\% reduction in parameters and a 90\% decrease in computational complexity compared to existing state-of-the-art methods, while maintaining comparable reconstruction accuracy. 

  \item We propose two pivotal designs within MLPHand: the light-weight Skeleton2Mesh model, and the multi-view geometry feature fusion prediction module. They are designed to ensure the model's efficiency and effectiveness.

  \item We perform extensive experiments to validate the effectiveness and efficiency of our proposed method. The results demonstrate that our method achieves state-of-the-art performance in terms of running time, while maintaining competitive performance in reconstruction accuracy when compared to strong baseline methods.
\end{itemize}


\section{Related Work}
\label{sec:relatedwork}
\subsection{Image-Based Hand Reconstruction}
\label{related_hand_recon}
Over the past several decades, monocular hand reconstruction has been at the forefront of research interest within the field. These methods can generally be divided into two categories: model-based methods and model-free methods. Model-based methods aim to regress the parameters of a hand parametric model, such as MANO~\cite{romero2022embodied}, by employing RGB cameras. 
Notable examples include approaches proposed in~\cite{wang2020rgb2hands,boukhayma20193d,hasson2019learning,kong2022identity,zimmermann2019freihand,zhou2020monocular}. On the other hand, model-free methods, such as those presented in~\cite{chen2021camera,ge20193d,kolotouros2019convolutional,kulon2020weakly,chen2022mobrecon}, take a different approach by directly regressing the hand surface. Despite receiving widespread attention, monocular methods inherently pose an ill-posed problem, making accurate reconstruction a challenge. 
In light of this, multi-view hand reconstruction emerges as a potential alternative. 
Hence, in recent, POEM~\cite{yang2023poem} and Spectral Graphormer~\cite{Tse_2023_ICCV} were proposed to reconstruct single-hand and two-hand from multi-view images, repectively.
While these approaches yield impressive reconstruction accuracy, they requires substantial computational resources for inference, making real-time performance difficult to achieve. 
In contrast, this paper shifts the focus towards real-time multi-view single hand reconstruction. 
To this end, we introduce a real-time method named MLPHand. 
It leverages an efficient and effective MLP-based skeleton-to-mesh strategy, enabling real-time applications.


\subsection{MLP-based 3D Modeling}
\label{sec:related_mlp}
Multi-layer Perceptrons (MLPs) have been successfully employed in a variety of applications, particularly in the domain of 3D shape and scene representation. Classic and recent works \cite{mescheder2019occupancy, fu2022geo, yu2022monosdf, park2019deepsdf, chou2023diffusion, mildenhall2021nerf, wang2022clip, hong2022headnerf, pumarola2021d, mihajlovic2022coap} have demonstrated the application of MLPs in modeling implicit functions for objects such as signed distance fields \cite{yu2022monosdf, park2019deepsdf, chen2022alignsdf}. These methods capitalize on the representative capability of MLPs to generate intricate, high-dimensional interpretations of 3D shapes, offering advantages over traditional techniques.
In particular, the use of MLPs for human hand modeling has emerged as a promising area of research. Innovations in this field include significant contributions such as HALO \cite{karunratanakul2021skeleton}, LISA \cite{corona2022lisa}, and COAP \cite{mihajlovic2022coap}. These advancements highlight the potential of MLP-based methods in capturing the complexity of human hand geometries.

Building upon these insights in human implicit geometry modeling \cite{karunratanakul2021skeleton, corona2022lisa, mihajlovic2022coap, chen2023hand, deng2020nasa}, MLPHand includes an MLP-based Skeleton2Mesh model recovering hand meshes from skeletons.
On the other hand, inspired by Zhang \textit{et.al.}~\cite{zhang2023adding}, we propose an MLP-based strategy for local geometry feature fusion that enhances the entire framework and still allows for real-time inference.

\begin{figure*}[!t]
    \centering
    \includegraphics[width=1.0\textwidth]{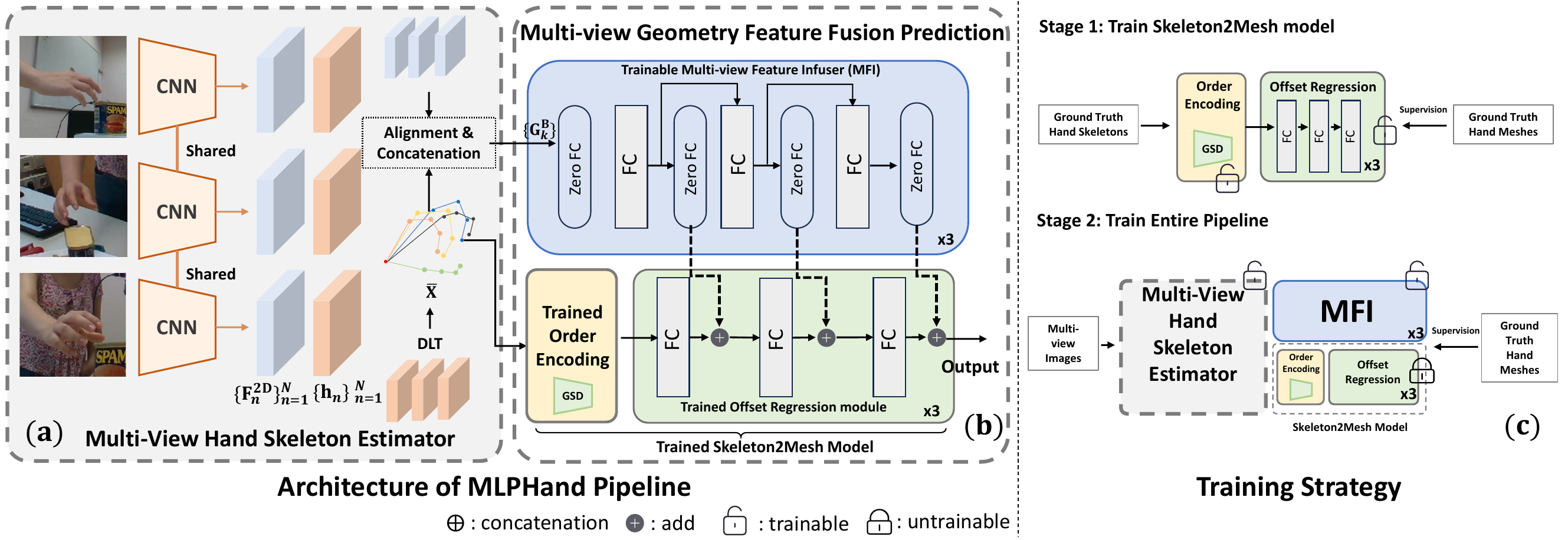}
    \caption{The Overview of MLPHand and its training strategy. MLPHand consists of: (${\bf a}$) multi-view hand skeleton estimator and (${\bf b}$) multi-view geometry feature fusion prediction module. (${\bf c}$) displays two training stage of our method.}
    \label{fig:enter-label}
\end{figure*}

\section{Method}
\label{sec:method}

\subsection{Overview}
\label{subsec:formulation}
MLPHand outputs the surface mesh, ${{\bf V} \in \mathbb{R}^{778 \times 3}}$ (MANO-style~\cite{romero2022embodied}), given $N$ RGB images ${\bf \mathcal{I}} = \left\{ {\bf I}_n \right\}_{n=1}^N$ acquired from different cameras with known positions and orientations.
It begins with a multi-view hand skeleton estimator (see \cref{fig:enter-label} (a)) to obtain a reference skeletons ${\bf \bar{X}} \in \mathbb{R}^{21 \times 3}$ and image feature maps ${\left\{{\bf F}^{2D}_n\right\}_{n=1}^N} \subset \mathbb{R}^{H \times W \times C}$. 
Next, both aligned and concatenated bone-level geometry features $\{{\bf G}^B_k\}_{k=1}^{20}$ and the reference skeleton ${\bf \bar{X}}$ are fed to the Multi-view Geometry Feature Fusion prediction (termed as MGFP for simplicity) module (see \cref{fig:enter-label} (b)) to obtain the hand meshes as final output. 
The MGFP contains a trained Skeleton2Mesh model achieving accurate and efficient mapping from hand skeletons to meshes, and Multi-view Feature Infuser (MFI) leveraging the visual features to enhance the accuracy of trained Skeleton2Mesh model.

As shown in \cref{fig:enter-label} (c), we take a two stage training procedure for better performance: first, we train the Skeleton2Mesh model (see \cref{fig:illustration} (a) and (b)) with ground truth pairs of hand skeletons and meshes; second, we train the entire pipeline with multi-view images and ground hand meshes. Next, we provide detailed introductions to these modules.
\subsection{Multi-View Hand Skeleton Estimator}
\label{subsec:pose_estimation}
Given the set of multi-view images $\mathbf{\mathcal{I}}=\{ \mathbf{I}_n\}_{n=1}^N$, the multi-view hand skeleton estimator predicts the reference skeleton $\mathbf{\bar{X}}$ and captures image feature maps, ${\left\{{\bf F}^{2D}_n\right\}_{n=1}^N}$, from different viewpoints.
Specifically, we commence by estimating the 2D locations of hand skeletons, denoted as ${\bf s}_n \in \mathbb{R}^{21 \times 2}$, in each view ${\bf I}_n$. Following this, we employ Direct Linear Transform (DLT) triangulation, as described in \cite{hartley2003multiple}, to elevate these 2D poses into the 3D realm. 
On the other hand, similar to the approach used in \cite{yang2023poem}, we take the 2D likelihood heatmap as 2D hand skeleton representation\cite{wei2016convolutional} and then apply \textit{soft-argmax} operation\cite{sun2018integral} on it for ${\bf s}_n$.
Formally, we have:
\begin{equation}
\begin{aligned}
& {\bf h}_n=\mathcal{F}_{\gamma}({\bf I}_n),\\
&{\bf s}_n={\rm soft\mbox{-}argmax}({\bf h}_n),\\
&{\bf \bar{X}}={\rm DLT}({\bf s}_{1 \sim N},{\bf K}_{1 \sim N},{\bf T}_{1 \sim N}),
\end{aligned}
\label{eq:stage1}
\end{equation}
where ${\bf h}_n \in \mathbb{R}^{21 \times H_{h} \times W_h}$ denotes the heatmap representation of ${\bf s}_n$, $\mathcal{F}{\gamma}$ represents the Convolutional Neural Network (CNN) backbone network with parameters $\gamma$, and ${\bf K}_n$ and ${\bf T}_n$ denote the intrinsic and extrinsic camera matrix for the \textit{n}-th camera, respectively. 
In this paper, we specifically adopt ResNet34~\cite{he2016deep}, ResNet18~\cite{he2016deep}, and MobileNetV2~\cite{sandler2018mobilenetv2}, and conduct detailed analysis to each of them.


\subsection{Skeleton2Mesh Model}
Our core idea is that accurate hand reconstruction can directly be obtained with precise mapping from hand skeletons to hand meshes. 
By ensuring this mapping both accurate and efficient, we sidestep complex geometry refinement operations such as the cross-set Transformer in POEM~\cite{yang2023poem}, enabling the whole multi-view hand reconstruction method to function in real-time.
Hence we propose Skeleton2Mesh model adopting the \textbf{Per-Bone reconstruction} strategy and the \textbf{Tri-Axis modeling} strategy for achieving this mapping.
Specifically, Per-Bone reconstruction divides the non-convex hand shape into 20 bone-wise convex mesh components with a defined order (shown in \cref{fig:illustration} (c)), and represents these geometries in a parameter-sharing manner making network light-weight.
Tri-Axis modeling employs three parallel MLP networks to independently regress the $xyz$ offsets relative to the midpoints of bones, which is an efficient modeling manner.
We design order encoding module (see \cref{fig:illustration} (a)) and offset regression module (see \cref{fig:illustration} (b)) for achieving Per-Bone reconstruction and Tri-Axis modeling respectively.

\subsubsection{Order Encoding.}
Given a hand skeleton ${\bf X} := \left\{ {\bf x}_k \in \mathbb{R}^3 \mid i=k,\dots,21 \right\}$, the order encoding module transforms these keypoints into bone-level features, 
which encapsulate both spatial and order information of the bones. 

\begin{figure*}[!t]
    \centering
    \includegraphics[width=2.0\columnwidth]{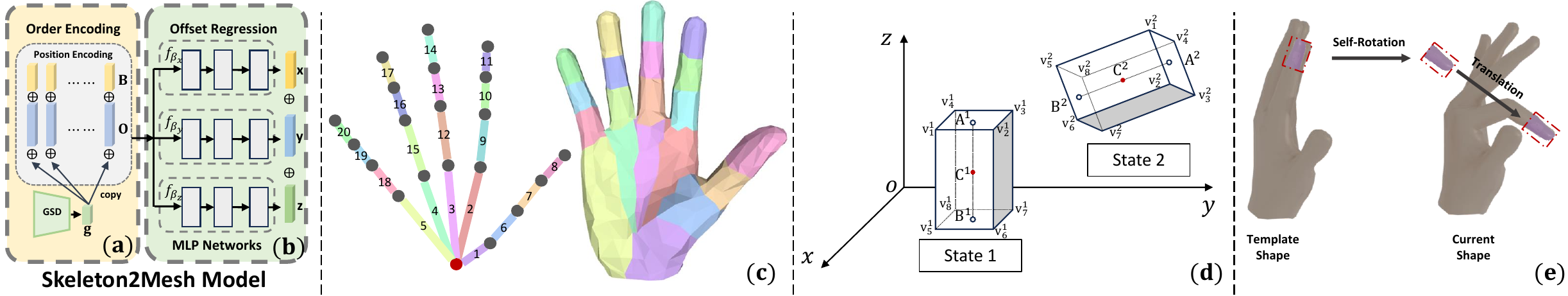}
    \caption{The order encoding module (a) and (b) offset regression module both belong to the Skeleton2Mesh model. (c) displays the skeleton-based convex decomposition of non-convex hand mesh (\textit{right}) and bone's order (\textit{left}). We place the detailed decomposition process in \textit{supplemental material}. (d) displays a toy example of cube transformation. (e) displays the bone-wise self-rotation and translation from the template to the current skeleton-aligned shape. }
    \label{fig:illustration}
\end{figure*}

Specifically, to capture per-bone spatial information effectively, we utilize the 3D coordinates of two endpoints of each bone as part of input, represented as ${\bf B} := \{ {\bf b}_k=[{\bf x}_p^k, {\bf x}_c^k] \in \mathbb{R}^6 $ 
$ \mid k=1,\dots,20 \}$, where $[\cdot]$ symbolizes vector concatenation, ${\bf x}_p^k$ and ${\bf x}_c^k$ denote the parent and child nodes of the $k$-th bone in the hand kinematic tree (illustrated in \cref{fig:illustration} (c)). 
Additionally, we introduce a learnbale Global Spatial Descriptor (GSD) ${\bf g} = {\rm MLP}({\bf X})\colon \mathbb{R}^{21 \times 3} \rightarrow \mathbb{R}^{100}$ to provide supplementary spatial information to prevent geometric collapse of the palm (see \cref{fig:gsd_ablation}).
We include derivation of obtaining 6D pose from ${\bf B}$ and the analysis of the GSD in our \textit{supplemental material}.

On the other hand, for capturing the order information, we utilize one-hot encoded vectors,
$\mathbf{O} := \{ \mathbf{o}_k \in \mathbb{R}^{20} \mid k=1,\dots,20 \}$,
as identifiers to reflect the order information that is defined in \cref{fig:illustration}(c).
Contrasting with previous implicit per-bone reconstruction models~\cite{deng2020nasa, corona2022lisa, chen2023hand, mihajlovic2022coap}, which model each local geometry with different networks, $\mathbf{O}$ enables a parameter-sharing manner for all per-bone reconstruction networks in the Skeleton2Mesh model. 
By doing so, we substantially reduce the number of parameters in the Skeleton2Mesh model, thereby enhancing its suitability for mobile deployments due to lower computational requirements. 
In summary, our order encoding operation is formulated as:
\begin{equation}
    {\rm OE}({\bf {X}}) := \left\{ \left[ {\rm PE}([{\bf {b}}_k, {\bf o}_k]), {\bf g} \right], k=1,\dots,20 \right\},
    \label{eq:orderencoding}
\end{equation}
where ${\rm PE}(\cdot)$ denotes the element-wise position encoding function that empirically enhances model performance.
The position encoding used in this context follows the formulation presented by Mildenhall \textit{et.al.}~\cite{mildenhall2021nerf}:
\begin{equation}
\begin{split}
    {\rm PE}(x) &= ( \sin(2^0 \pi x), \cos(2^0 \pi x), \ldots, \sin(2^{L-1} \pi x), \\
    &\cos(2^{L-1} \pi x) ).
    \label{eq:pe}
\end{split}
\end{equation}
Here, $L$ represents the level of frequency bands used in encoding, where $L=2$ for elements in ${\bf o}_k$ and $L=5$ for elements in ${\bf b}_k$. Therefore, ${\rm PE}(\cdot)\colon \mathbb{R}^{26} \rightarrow \mathbb{R}^{140}$, and consequently, ${\rm OE}({\bf {X}}) \in \mathbb{R}^{20 \times 240}$. For simplicity, we refer to the $k$-th row of ${\rm OE}({\bf {X}})$ as ${\rm OE}({\bf {X}})_k = [{\rm PE}([{\bf {b}}_k, {\bf o}_k]), {\bf g}]$. 
By way of \cref{eq:orderencoding}, ${\rm OE}({\bf {X}})_k$ integrates both spatial and order information of the $k$-th bone.
The effectiveness of ${\rm PE}(\cdot)$ and GSD is demonstrated in \cref{sec:ablation}.

\subsubsection{Offset Regression.}
With encoded bone-level feature from order encoding module, another offset regression module is proposed to regress the $xyz$ coordinates of the vertices of individual per-bone meshes from these features. 

On the one hand, the design of offset regression module is inspired by an observation on the rigid object transformation:
As depicted in \cref{fig:illustration}(d), the transformation of the cube from state 1 to state 2 can be divided into two distinct operations: first, a self-rotation to attain a specified orientation; second, a translation to reach the specified position.
Let us denote the vertices of the cube in state 1 and state 2 as $\{ {\bf v}_n^1 \}_{n=1}^8 \subset \mathbb{R}^3$ and $\{ {\bf v}_n^2\}_{n=1}^8 \subset \mathbb{R}^3$, respectively.
Considering a rotation matrix ${\bf R}^T = ({\bf r}_1, {\bf r}_2, {\bf r}_3)$, where ${\bf r}_n \in \mathbb{R}^{3}$ for $n \in \{1,2,3\}$. The self-rotation is given by:
\begin{equation}
\begin{aligned}
({\bf v}_1^2, \dots, {\bf v}_8^2) &= {\bf R}({\bf v}_1^1, \dots, {\bf v}_8^1) \\
&=\left( \begin{matrix}
{\bf r}_1^T \cdot ({\bf v}_1^1, \dots, {\bf v}_8^1) \\
{\bf r}_2^T \cdot ({\bf v}_1^1, \dots, {\bf v}_8^1) \\
{\bf r}_3^T \cdot ({\bf v}_1^1, \dots, {\bf v}_8^1)
\end{matrix} \right)
\end{aligned}
\label{eq:triaxis}
\end{equation}
This indicates that for self-rotation transformations, each coordinate can be treated by itself, and we call this idea \textit{coordinate-separable}.

On the other hand, as shown in \cref{fig:illustration}(e), the reconstruction process of one local mesh is also delineated into two phases: the self-rotation dictated by the current bone orientation and the translation determined by the bone's midpoint. 
This means the reconstruction process of per-bone in our assumption is similar with the cube transformation in \cref{fig:illustration}(d). 
Hence, taking into account the Tri-Axis observation in \cref{eq:triaxis} and the nonlinear nature of hand muscle dynamics, we employ three separate MLP networks with \textit{LeakyReLU} activation (excluding the last layer) to regress these bone-wise offsets. The mathematical framework for the reconstructions of per-bone meshes is given as follows: for the $k$-th bone, considering the hand skeleton ${\bf X}$, we have
\begin{equation}
\begin{split}
f_{\beta}({\rm OE}({\bf {X}})_k) = \left( \begin{matrix}
{\rm MLP}_{{\beta}_x}({\rm OE}({\bf {X}})_k) \\
{\rm MLP}_{{\beta}_y}({\rm OE}({\bf {X}})_k) \\
{\rm MLP}_{{\beta}_z}({\rm OE}({\bf {X}})_k)
\end{matrix} \right) \in \mathbb{R}^{100 \times 3},
\end{split}
\label{eq:function_gamma2}
\end{equation}
where $\beta = \{ {\beta}_x, {\beta}_y, {\beta}_z \}$ represents the parameters of the coordinate-wise MLPs. Each MLP in \cref{eq:function_gamma2} outputs a 100-dimensional vector, and the final vertex coordinates are obtained by element-wise concatenating the output vectors and trimming the excess.
Due to space constraints, we place detailed \textbf{loss function} setting and additional implementation features of the Skeleton2Mesh model in the \textit{supplemental material}.

In \cref{sec:ablation}, our ablation analysis validates that this modeling methodology is notably effective and efficient, with just three layers in each ${\rm MLP}_{{\beta}_i}(\cdot)$, for $i \in \{ x, y, z \}$, being sufficient to attain accurate hand reconstructions.

\begin{figure*}[!t]
    \centering \includegraphics[width=2.0\columnwidth]{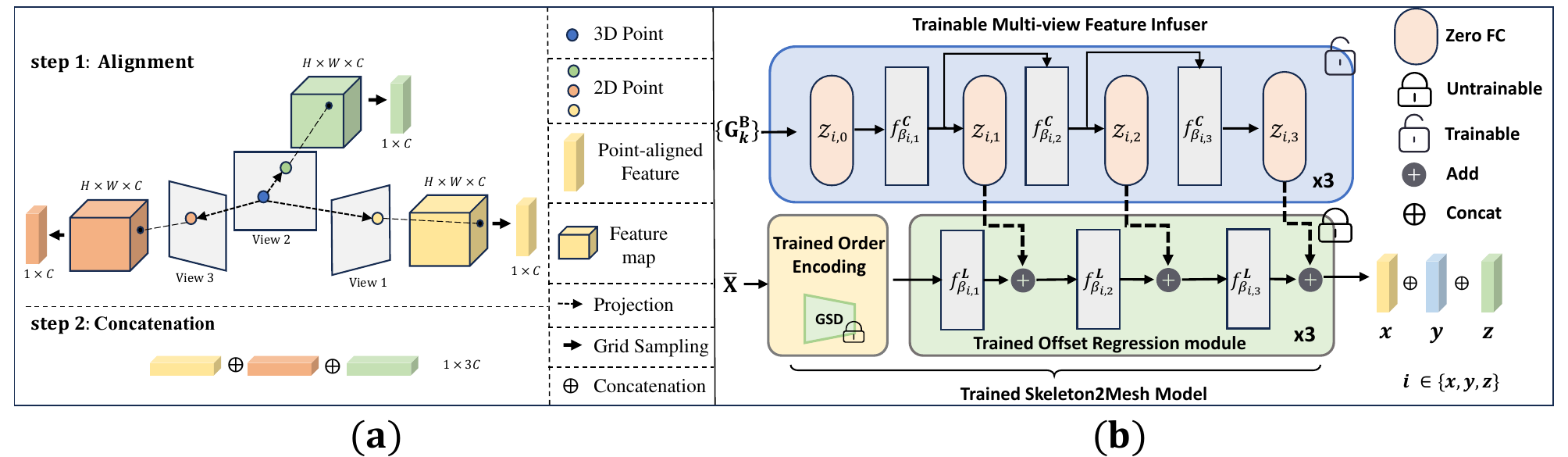}
    \caption{(a) depicts the alignment and concatenation operations applied to a given 3D point during the cross-view geometry feature fusion process (3 views for simplicity). (b) displays the feature forward-propagation process in MGFP module, during the reconstruction process of the \textit{k}-th bone's local mesh.}
    \label{fig:illustration2}
\end{figure*}

\subsection{Multi-View Geometry Feature Fusion Prediction}
\label{subsec:vision_enhance_predict}
A system that simply cascades the multi-view hand skeleton estimator with the trained Skeleton2Mesh model can predict hand meshes from multi-view images in real time, as experimented in \cref{tab:ablation1}.
However, this straightforward pipeline neglects the visual information present in images, which is critical for capturing the details of hand shapes.
Inspired by recent advancement in conditional control of trained models~\cite{zhang2023adding}, we propose MGFP module (see \cref{fig:enter-label}(b)) to fully exploit the visual features to enhance the performance of trained Skeleton2Mesh model.

Remember that we also obtain visual feature maps $\left\{ {\bf F}_n^{\text{2D}} \right\}_{n=1}^N$, while predicting 2D hand poses from multi-view images.
To extract hand-related geometry features from $\left\{ {\bf F}_n^{\text{2D}} \right\}_{n=1}^N$, we reproject the 3D points ${\bf \bar{X}}$ onto image planes utilizing intrinsic camera matrices $\left\{{\bf K}_n\right\}_{n=1}^N$ and extrinsic camera matrices $\left\{{\bf T}_n\right\}_{n=1}^N$ to obtain reprojected 2D hand skeletons $\left\{{\bf P}^{\text{2D}}_n\right\}_{n=1}^N \subset \mathbb{R}^{21 \times 2}$.
We then align $\left\{{\bf P}^{\text{2D}}_n\right\}_{n=1}^N$ with $\left\{ {\bf F}_n^{\text{2D}} \right\}_{n=1}^N$ via grid sampling~\cite{saito2019pifu}, and perform channel-wise concatenation of the same keypoint-class feature across different views to obtain keypoint-level multi-view geometry features $\left\{{\bf G}^K_k \right\}_{k=1}^{21} \subset \mathbb{R}^{NC}$.
The above process for each 3D point in ${\bf \bar{X}}$ is displayed in \cref{fig:illustration2}(a).
Subsequently, bone-level multi-view geometry features $\left\{{\bf G}^B_k\right\}_{k=1}^{20}$, where ${\bf G}^B_k \in \mathbb{R}^{2NC}$, are derived by channel-wise concatenation along the hand kinematic tree illustrated in \cref{fig:illustration}(a).
In this way, $\left\{{\bf G}^B_k\right\}_{k=1}^{20}$ fuses all bone-related visual features from different views.

Next, we put the set of bone-level geometry features $\{{\bf G}^B_k\}_{k=1}^{20}$ into our MGFP module, as showcased in \cref{fig:enter-label}(d). 
This module infuses $\{{\bf G}^B_k\}_{k=1}^{20}$ into trained Skeleton2Mesh model to enhance its geometry accuracy via the Multi-view Feature Infuser (termed as MFI) module. 
The MFI module comprises trainable copy of the offset regression module in trained Skeleton2Mesh model and  "Zero FC" (a fully-connected layer initialized with zeros) layers. 
The "Zero FC" is designed in the spirit of \cite{zhang2023adding} to establish parameter growth from a neutral base, thus mitigating introduction of extraneous noise that might otherwise impair the performance of the trained Skeleton2Mesh model.
As shown in the detailed step-by-step fusion procedure in ~\cref{fig:illustration2}(b), given $\mathbf{G}_k^B$ and the hand skeleton estimation $\mathbf{\bar{X}}$, the fusion process for the $k$-th bone can be formulated as:
\begin{equation}
\begin{aligned}
\left\{
\begin{array}{ll}
{\bf z}_{i,0}=\mathcal{Z}_{i,0}({\bf G}_k^B), & \\
{\bf z}_{i,j}=\mathcal{Z}_{i,j}(f_{{\beta}_{i,j}}^{\bf C}({\bf z}_{i,j-1})), & \text{for } j=1,2,3,\\
{\bf e}_{i,0}={\rm OE}({\bf \bar{X}})_k, & \\
{\bf e}_{i,j}={\bf z}_{i,j} + f_{{\beta}_{i,j}}^{\bf L}({\bf e}_{i,j-1}), & \text{for } j=1,2,3,\\
\end{array}
\right.
\end{aligned}
\label{eq:stage3}
\end{equation}
where $i \in \{x,y,z\}$ denotes the spatial axis, $\mathcal{Z}_{i,j}$ (for $j \in \{0,1,2,3\}$) refers to one ``Zero FC'' layer, $f_{{\beta}_{i,j}}^{\bf L}$ signifies the locked $j$-th MLP layer (parameters frozen) for the $i$-th axis in the Skeleton2Mesh model, ${\bf z}_{i,j}$ is the geometry-enhancement feature which is extracted via "Zero FC" and infused into the trained Skeleton2Mesh model via layer-wise vector addition, and $f_{{\beta}_{i,j}}^{\bf C}$ denotes its trainable counterpart. The output for the $k$-th bone, $[{\bf e}_{x,3},{\bf e}_{y,3},{\bf e}_{z,3}]$, is belong to $\mathbb{R}^{100 \times 3}$. Finally by discarding superfluous points, we obtain the final vertices of the k-th bone's mesh.
Note that \textbf{loss function} setting in stage 2 is in our \textit{supplemental material}. 

\subsection{Training Strategy}
By training the Skeleton2Mesh model on ground truth pairs of hand skeletons and meshes prior to end-to-end training, this strategy enables the model to learn prior knowledge from hand skeletons to meshes.  
With this learned prior knowledge, our model achieves a kind of posterior estimation from multi-view images and obtains better accuracy. 
We provide additional experimental results in our \textit{supplemental material} to further demonstrate the effectiveness of this strategy.

\section{Experiments}
\subsection{Implementation Details}
Our models are implemented using the PyTorch framework. We utilize the Adam optimizer~\cite{kingma2014adam} with a mini-batch size of 32 for training the network. The images resolution employed in this work is 256 × 256. We initialize all CNN backbones used in our experiments with ImageNet~\cite{russakovsky2015imagenet} pretrained weights. To achieve optimal performance, we adopt a two-stage training strategy. In the first stage, we focus on training the Skeleton2Mesh model. We set the initial learning rate to 1e-4 and train this module for 300 epochs. And we reduce the learning rate by half every 50 epochs. Once the Skeleton2Mesh model has been adequately trained, we freeze its parameters. In the second stage, we proceed to train the entire pipeline. The training duration for this stage is set to 100 epochs, with an initial learning rate of 1e-4. To further optimize the learning process, we decrease the learning rate by a factor of 10 after the 70th epoch.

\subsection{Datasets}
We conduct our experiments on three prominent multi-view human hand reconstruction datasets, namely DexYCB-MV~\cite{chao2021dexycb}, HO3DV3-MV~\cite{hampali2021ho}, and OakInk-MV~\cite{yang2022oakink}, each with distinct characteristics.

\noindent \textbf{DexYCB-MV}~\cite{chao2021dexycb} is a rich dataset, encompassing 582K images, having 8 camera perspectives. To provide a fair basis for comparison with POEM~\cite{yang2023poem}, we adapt same data configuration with it, which comprises 25,387 multi-view (m.v.) frames, equivalent to 203,096 monocular frames, for training. Additionally, it includes 1,412 m.v. frames for validation, and 4,951 for testing.

\noindent \textbf{HO3D V3-MV}~\cite{hampali2021ho} has 103,462 images capturing hand-object interactions from up to 5 camera viewpoints. We adopt same data configuration with POEM. This results in a total of 9,087 m.v. frames (or 45,435 monocular frames) for the training set and 2,706 m.v. frames for testing.

\noindent \textbf{OakInk-MV}~\cite{yang2022oakink} is a robust dataset containing 230K images derived from 4 camera observations. We adopt same data configuration with POEM. The OakInk-MV dataset comprises 58,692 m.v. frames (equivalent to 234,768 monocular frames) in the training set, and 19,909 m.v. frames in the testing set.

\noindent\textbf{FreiHAND}~\cite{zimmermann2019freihand} is a \textbf{monocular} hand reconstruction dataset, which contains 32,560 hand meshes for training and 3960 for testing. We conduct the ablation about the design of Skeleton2Mesh model on it.


\subsection{Evaluation Metrics}
We present the \textbf{MPJPE} and \textbf{MPVPE} (in millimeters), which stand for mean per keypoint (joint) and per vertex position error, respectively. 
It is worth noting that, as MANO~\cite{romero2022embodied} offers a pre-trained mapping from \textbf{V} to \textbf{X}, the reported keypoints \textbf{X} are a by-product of the final vertex \textbf{V} from the second stage. 
Additionally, we also evaluate the MPVPE and MPVPE within a root-relative (RR) system and under Procrustes Analysis (PA)~\cite{gower1975generalized}.
To compare the efficiency, we also report \textbf{Multi-Adds} (in Giga), \textbf{Params} (in Million) and \textbf{FPS}, which stand for the count of multiply-add operations, the number of parameters and frames of per second on NVIDIA GeForce RTX 3090 GPU, respectively.

\begin{table*}[!t]
\centering
\scalebox{0.9}{
    \begin{tabular}{c|c|c|c|c|c|c|c|c|c|c|c}
    \toprule
        Dataset & Method & MPVPE$\downarrow$ & RR-V$\downarrow$ & PA-V$\downarrow$ & MPJPE$\downarrow$ & RR-J$\downarrow$ & PA-J$\downarrow$ & Backbone & Params$\downarrow$ & Multi-Adds$\downarrow$ & FPS$\uparrow$ \\ \midrule \hline
        \multirow{5}*{\rotatebox{90}{Oaklnk-MV}} & MVP\cite{zhang2021direct} & 9.69 & 11.75 & 7.74 & 7.32 & 9.99 & 4.97 & ResNet34 & 53.57M & 21.27G & 22\\ \cline{2-12}
        ~ & POEM\cite{yang2023poem} & \underline{6.2} & \underline{7.63} & \textbf{4.21} & \underline{6.01} & \textbf{7.46} & \textbf{4.00} & ResNet34 & 40.49M & 31.35G & 19\\ \cline{2-12}
        ~ & \multirow{3}*{Ours} & \textbf{6.17} & \textbf{7.49} & \underline{4.26} & \textbf{6.00} & \underline{7.52} & \underline{4.09} & ResNet34 & 26.24M & 24.64G & 57 \\ 
        ~ & ~ & 6.53 & 7.86 & 4.49 & 6.37 & 7.89 & 4.33 & ResNet18 & \underline{16.13M} &\underline{ 14.97G} & \textbf{71} \\ 
        ~ & ~ & 6.61 & 7.93 & 4.59 & 6.43 & 7.95 & 4.41 & MobileNetv2 & \textbf{11.58M} & \textbf{3.09G} & \underline{65}\\ \hline
        \midrule 
        \multirow{5}*{\rotatebox{90}{DexYCB-MV}} & MVP\cite{zhang2021direct} & 9.77 & 12.18 & 8.14 & \underline{6.23} & 9.47 & 4.26 & ResNet34 & 57.24M & 42.53G & 21 \\ \cline{2-12}
        ~ & POEM\cite{yang2023poem} & \textbf{6.13} & \textbf{7.21} & \textbf{4.00} & \textbf{6.06} & \textbf{7.30} & \textbf{3.93} & ResNet34 & 40.49M &  62.68G & 16 \\ \cline{2-12}
        ~ & \multirow{3}*{Ours} & \underline{6.20} & \underline{7.30} & \underline{4.15} & \underline{6.23} & \underline{7.41} & \underline{4.11} & ResNet34 & 26.98M & 49.19G & 54 \\ 
        ~ & ~ & 6.69 & 7.8 & 4.39 & 6.65 & 7.91 & 4.34 & ResNet18 & \underline{16.87M} & \underline{29.86G} & \textbf{65} \\ 
        ~ & ~ & 6.87 & 7.91 & 4.54 & 6.84 & 8.03 & 4.49 & MobileNetv2 & \textbf{12.50M} & \textbf{6.14G} & \underline{63} \\ \hline
        \midrule 
        \multirow{5}*{\rotatebox{90}{HO3D-MV}} & MVP\cite{zhang2021direct} & 20.95 & 27.08 & \underline{10.04} & 18.72 & 24.9 & 10.44 & ResNet34 & 54.49M & 26.59G & 22\\ \cline{2-12}
        ~ & POEM\cite{yang2023poem} & \textbf{17.2} & \textbf{21.45} & \textbf{9.97} & \textbf{17.28} & \textbf{21.94} & \textbf{9.6} & ResNet34 & 40.49M & 39.18G& 18\\ \cline{2-12}
        ~ & \multirow{3}*{Ours} & \underline{18.69} & \underline{23.28} & 10.54 & \underline{18.7} & \underline{23.76} & \underline{10.11} & ResNet34 &  26.42M& 30.74G & 58 \\ 
        ~ & ~ & 20.69 & 25,56 & 11.91 & 20.61 & 26.09 & 11.42 & ResNet18 & \underline{16.32M}  & \underline{18.66G} &\textbf{69}\\ 
        ~ & ~ & 20.7 & 27 & 11.76 & 20.7 & 26.3 & 11.3 & MobileNetv2 &\textbf{11.81M}& \textbf{3.80G} & \underline{64}\\ \hline
    \bottomrule
    \multicolumn{8}{l}{\textbf{Best}; \underline{Second best}.}
    \end{tabular}}
    \vspace{2mm}
\caption{The quantitative results (mm) of evaluations on Oaklnk-MV, DexYCB-MV and HO3D-MV. \textbf{Note} that although the MobileNetV2 backbone does not achieve the fastest inference speed, despite having the lowest Multi-ADD operations, this is because its depth-wise convolution operation incurs a high amount of memory access and is more compatible with CPU devices than GPU devices.}
\label{tab:quanti}
\end{table*}

\subsection{Results and Comparisons}
We compared our method with MVP~\cite{zhang2021direct} and POEM~\cite{yang2023poem} across three different datasets: Oaklnk-MV, DexYCB-MV, and HO3D-MV. The results, as indicated in Table \ref{tab:quanti}, demonstrate the superiority of our method in several aspects.

\begin{figure}[!h]
    \centering
    \includegraphics[width=1.0\columnwidth]{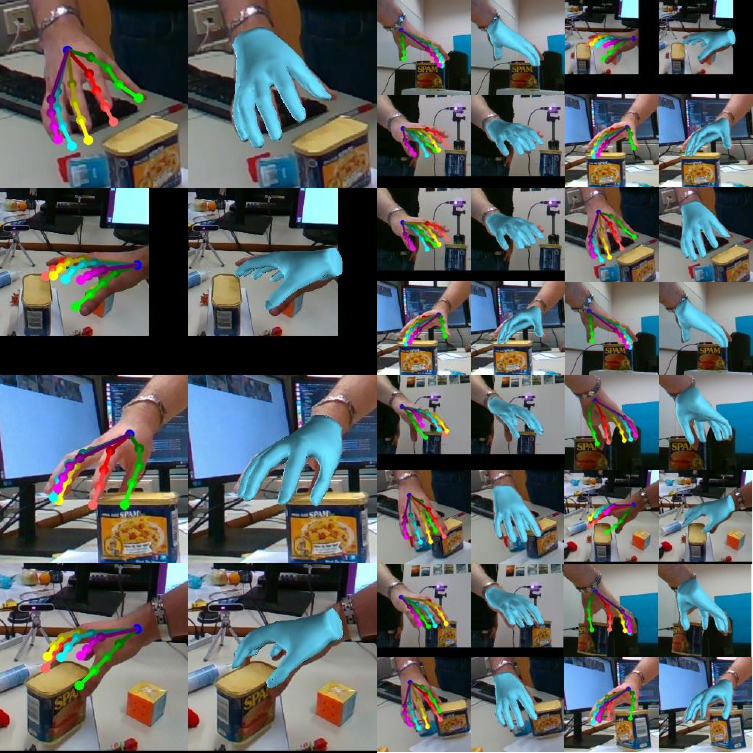}
    \caption{The qualitative display of the DexYCB-MV dataset.}
    \label{fig:quali_dexycb_part}
\end{figure}

On Oaklnk-MV Dataset, in terms of efficiency, our method significantly outperforms both MVP and POEM. Notably, using ResNet34 as the backbone, our method achieves an FPS of 57, which is approximately three times faster than POEM (19 FPS) and MVP (22 FPS). We also witness a notable reduction in both parameters (26.24M compared to POEM's 40.49M and MVP's 53.57M) and Multi-Adds (24.64G compared to POEM's 31.35G).  When we further optimize our method using lighter backbones, such as ResNet18 and MobileNetv2, the advantages become more prominent. With ResNet18, our method achieves a real-time performance of 71 FPS, reducing the parameters to 16.13M and Multi-Adds to 14.97G. With MobileNetv2, our method achieves 65 FPS with the lowest parameters (11.58M) and Multi-Adds (3.09G) among all methods.  In terms of accuracy metrics (MPVPE, RR-V, PA-V, MPJPE, RR-J, PA-J), our method outperforms MVP and is comparable to POEM, showing that our method maintains competitive performance while being more efficient. Similar trends can be observed in the DexYCB-MV and HO3D-MV datasets. Our method consistently achieves higher FPS, and lower parameters and Multi-Adds compared to baselines. Meanwhile, the accuracy metrics of our method are generally comparable to the best performing ones, demonstrating the effectiveness of our approach in different scenarios. In conclusion, our method delivers comparable hand reconstruction results compared to existing methods while maintaining a significantly higher speed. The results provide solid evidence that our method can achieve real-time performance without compromising accuracy, making it an ideal solution for real-world applications that require both efficiency and effectiveness.
In \cref{fig:quali_dexycb_part}, we visualize some qualitative results of the DexYCB-MV dataset.
Specifically, for each multi-view frame, we draw its result from 5 different views. One of the views is in normal size and the other four views are half size. 
We place full \textbf{qualitative results} in our \textit{Supplemental material}, due to the constraint of pages limitation.

\subsection{Performance of Skeleton-to-mesh Modeling}
Table~\ref{tab:mlphand_analysis} presents a comparison of computational complexity between our method and POEM, specifically excluding the 3D hand skeleton estimation process. For a consistent evaluation, this experiment was conducted on the Oaklnk-MV dataset using ResNet-34 as the shared backbone architecture. This ensures that both MLPHand and POEM utilize the same process for skeleton estimation, facilitating a fair comparison of their computational demands.
As reported in \cref{tab:mlphand_analysis}, it is evident that our approach significantly outperforms POEM in terms of computational efficiency and model size. Specifically, our method reduces the Multi-Adds from POEM's 6.76G to a mere 0.05G, representing an impressive reduction of over 99\%. This drastic reduction in computational complexity highlights the efficiency of our method, making it suitable for applications with stringent computational constraints. In terms of model size (Params), our method also shows a notable advantage, shrinking the number of parameters from POEM's 18.70M to just 2.12M. This represents an approximately 89\% reduction, underscoring the compact nature of our model. Turning our attention to the accuracy metrics, our method (Ref MPJPE, MPVPE, MPJPE) outperforms POEM across all three metrics. Our method achieves a Ref MPJPE of 7.32, MPVPE of 6.17, and MPJPE of 6.00, all of which are lower than the corresponding values of POEM. This demonstrates that our method's accuracy is not compromised despite its significant reduction in computational complexity and model size. These improvements can be attributed to MGFP module. By leveraging multi-view visual features, this module converts a human hand skeleton into a mesh more effectively and efficiently, making our approach superior to POEM.

\begin{table}[!ht]
    \centering
    \scalebox{0.7}{
    \begin{tabular}{c|c|c|c|c|c}
    \toprule
    
        Method & Multi-Adds$\downarrow$ & Params$\downarrow$ & Ref MPJPE$\downarrow$ & MPVPE$\downarrow$  & MPJPE$\downarrow$ \\ \hline
        POEM & 6.76G & 18.70M & 7.36 & 6.2 & 6.01 \\ \hline
        Ours & \textbf{0.05G} & \textbf{2.12M} & \textbf{7.32} & \textbf{6.17} & \textbf{6.00} \\ 
        \bottomrule
    \end{tabular}}
    \caption{The quantitative analysis of the skeleton-to-mesh process. Experiment is conducted on Oaklnk-MV, and the backbone is ResNet-34. Ref MPJPE indicates the MPJPE of the reference skeleton, which is ${\bf \Bar{X}}$ in MLPHand.}
\label{tab:mlphand_analysis}
\end{table}

\subsection{Robustness Testing and Future Work}
\label{sec:robustness}
In Table \ref{tab:robustnesstest}, we demonstrate the robustness of our trained Skeleton2Mesh model in the presence of noise. To conduct this test, we inject Gaussian Noise, denoted as $\mathcal{N}(0, \sigma^2)$, into ground-truth skeletons ${\bf X}$ as input. We vary the noise level by adjusting the variance $\sigma^2$ from 0 to 20. The results of this robustness test reveal a positive correlation between the accuracy of the input skeleton and the performance of this module. As the input skeleton's accuracy improves (i.e., as the noise level decreases), the performance of this module correspondingly enhances. This observation suggests that the performance of hand reconstruction can be substantially boosted by designing a more robust pose estimator. In essence, the quality of the input skeleton significantly influences the efficacy of the Skeleton2Mesh model, highlighting the importance of robust pose estimation in improving hand reconstruction performance.
Additionally, this experiment further showcases the potential of our two-stage MLPHand framework, in which the trained Skeleton2Mesh model is compatible with various skeleton estimation systems, including monocular, non-monocular, vision-based, and depth-based systems. 
Therefore, enhancing this framework by creating a more robust pose estimator and extending its applicability to more affordable data formats both represent promising avenues for future research.
\begin{table}[!h]
    \centering
    \scalebox{0.7}{\begin{tabular}{c|ccccc|c}
    \toprule
        Dataset & $\sigma^2=0$ & $\sigma^2=5$ & $\sigma^2=10$ & $\sigma^2=15$ & $\sigma^2=20$ & ~ \\ \midrule \hline
        \multirow{2}*{\thead{DexYCB\\-MV}}  & \textbf{0} & 3.56 & 5.04 & 6.18 & 7.14 & Ref MJ$\downarrow$ \\ 
        ~ & \textbf{0.6} & 3.27 & 4.58 & 5.59 & 6.44 & MV$\downarrow$ \\ \midrule \hline
        \multirow{2}*{\thead{Oaklnk\\-MV}} & \textbf{0} & 3.56 & 5.04 & 6.17 & 7.13 & Ref MJ$\downarrow$ \\ 
        ~ & \textbf{0.8} & 3.39 & 4.69 & 5.69 & 6.55 & MV$\downarrow$ \\ 
        \midrule \hline
        \multirow{2}*{\thead{HO3Dv3\\-MV}} & \textbf{0} & 3.56 & 5.05 & 6.18 & 7.14 & Ref MJ$\downarrow$ \\ 
        ~ & \textbf{16.38} & 17.7 & 18.57 & 19.24 & 19.83 & MV$\downarrow$ \\ 
        \bottomrule
    \end{tabular}}
    \caption{Robustness test of the MLP-based hand reconstruction module. Ref MJ indicates the MPJPE of reference skeletons injected with noise. MV indicates the MPVPE of generated meshes.}
    \label{tab:robustnesstest}
\end{table}

\begin{table}[!ht]
    \centering
    \begin{minipage}{.48\textwidth}
        \centering
        \scalebox{0.6}{
        \begin{tabular}{c|c|c|c|c|c|c}
        \hline
        Dataset & Full Model & Backbone & MPVPE$\downarrow$ & PA-V$\downarrow$ & MPJPE$\downarrow$ & PA-J$\downarrow$ \\ \hline
        \multirow{6}*{\rotatebox{90}{Oaklnk-MV}} & $\surd$ & ResNet34 & \textbf{6.17} & \textbf{4.26} & \textbf{6.00} & \textbf{4.09} \\ 
        ~ & $\times$ & ResNet34 & 6.59 & 4.59 & 6.43 & 4.41 \\ \cline{2-7}
        ~ & $\surd$ & ResNet18 & \textbf{6.53} & \textbf{4.49} & \textbf{6.37} & \textbf{4.33} \\ 
        ~ & $\times$ & ResNet18 & 6.83 & 4.72 & 6.67 & 4.54 \\ \cline{2-7}
        ~ & $\surd$ & MobileNetV2 & \textbf{6.61} & \textbf{4.59} & \textbf{6.43} & \textbf{4.41} \\ 
        ~ & $\times$ & MobileNetV2 & 6.88 & 4.86 & 6.70 & 4.64 \\ \hline

        \multirow{6}*{\rotatebox{90}{DexYCB-MV}}  & $\surd$ & ResNet34 & \textbf{6.28} & \textbf{4.15} & \textbf{6.23} & \textbf{4.11} \\ 
        ~ & $\times$ & ResNet34 & 6.93 & 4.68 & 6.92 & 4.65 \\ \cline{2-7}
        ~ & $\surd$ & ResNet18 & \textbf{6.69} & \textbf{4.39} & \textbf{6.65} & \textbf{4.34} \\ 
        ~ & $\times$ & ResNet18 & 7.19 & 4.78 & 7.18 & 4.74 \\ \cline{2-7}
        ~ & $\surd$ & MobileNetV2 & \textbf{6.87} & \textbf{4.54} & \textbf{6.84} & \textbf{4.49} \\ 
        ~ & $\times$ & MobileNetV2 & 7.4 & 4.93 & 7.39 & 4.87 \\ \hline

        \multirow{6}*{\rotatebox{90}{HO3Dv3-MV}} & $\surd$ & ResNet34 & \textbf{18.69} & \textbf{10.54} & \textbf{18.7} & \textbf{10.11} \\ 
        ~ & $\times$ & ResNet34 & 19.12 & 11.72 & 19.13 & 11.26 \\ \cline{2-7}
        ~ & $\surd$ & ResNet18 & \textbf{20.69} & \textbf{11.91} & \textbf{20.61} & \textbf{11.42} \\ 
        ~ & $\times$ & ResNet18 & 21.11 & 12.15 & 21.33 & 11.8 \\ \cline{2-7}
        ~ & $\surd$ & MobileNetV2 &\textbf{ 20.7} & \textbf{11.76} & \textbf{20.7} & \textbf{11.3} \\ 
        ~ & $\times$ & MobileNetV2 & 21.8 & 12.49 & 21.85 & 12.12 \\ \hline
        \end{tabular}
        }
        \vspace{5mm}
        \caption{The ablation study of the multi-view geometry feature fusion strategy. $\times$ indicates an experiment baseline with only the trained Skeleton2Mesh model.}
        \label{tab:ablation1}
    \end{minipage}\hspace{0.04\textwidth}%
    \begin{minipage}{.48\textwidth}
        \centering
        \scalebox{0.8}{
        \begin{tabular}{c|c|c|c|c|c|c|c}
        \hline
        Layers & PB & PE & GSD & PA-V$\downarrow$ & MV$\downarrow$ & Params$\downarrow$ &MA$\downarrow$ \\ \hline
        5 & $\surd$ & $\surd$ & $\surd$& 1.28 & 1.30& 0.90M &17.13M \\ \cline{2-8}
        4 & $\surd$ & $\surd$ & $\surd$ & 1.33 & 1.34& 0.70M & 13.18M\\ \cline{2-8}
        3 & $\surd$ & $\surd$ & $\surd$& 1.34 & 1.36 & 0.50M & 9.23M\\ \cline{2-8}
        3 & $\times$ & $\surd$ & $\surd$ & \textbf{1.21} & \textbf{1.23} &  7.23M & 7.23M\\ \cline{2-8}
        3 & $\surd$ & $\times$ & $\surd$& 1.38 & 1.39 & 0.41M & 7.48M\\ \cline{2-8}
        3 & $\surd$ & $\surd$ & $\times$ & 6.81 & 6.78 & 0.38M & 7.66M\\ \cline{2-8}
        2 & $\surd$ & $\surd$ & $\surd$ & 1.70 & 1.71 & \textbf{0.30M}& \textbf{5.29M}\\ \hline
        \end{tabular}
        }
        \vspace{5mm}
        \caption{The experiments of ablation study, where PB, MJ, MV and MA represent the per-bone reconstruction, MEJPE, MEVPE and Multi-ADDs (in Mega), respectively. All experiments is conducted on the FreiHand dataset.}
        \label{tab:ablation_basic_mlphand}
    \end{minipage}
\end{table}

\subsection{Ablation Study}
\label{sec:ablation}
\noindent{\bf The impact of the Multi-view Geometry Feature Fusion Strategy.}
In this part, we conduct  ablation study on the MGFP module. The aim of this study is to verify the effectiveness and contribution of our multi-view geometry feature fusion strategy to the overall performance of our model. The results are shown in Table \ref{tab:ablation1}.  For each combination of dataset and backbone architecture, we perform two sets of experiments: one with the MGFP module (indicated by $\surd$ in ``Full Model" column), and one with only trained Skeleton2Mesh model (indicated by $\times$ in ``Full Model" column). Across all datasets and backbone architectures, the full model consistently obtains better performance. It turns out that our multi-view geometry feature fusion strategy successfully infuses useful visual features into the trained Skeleton2Mesh model, such enhances its performance.

\begin{figure}[!h]
    \centering
    \includegraphics[width=1.0\columnwidth]{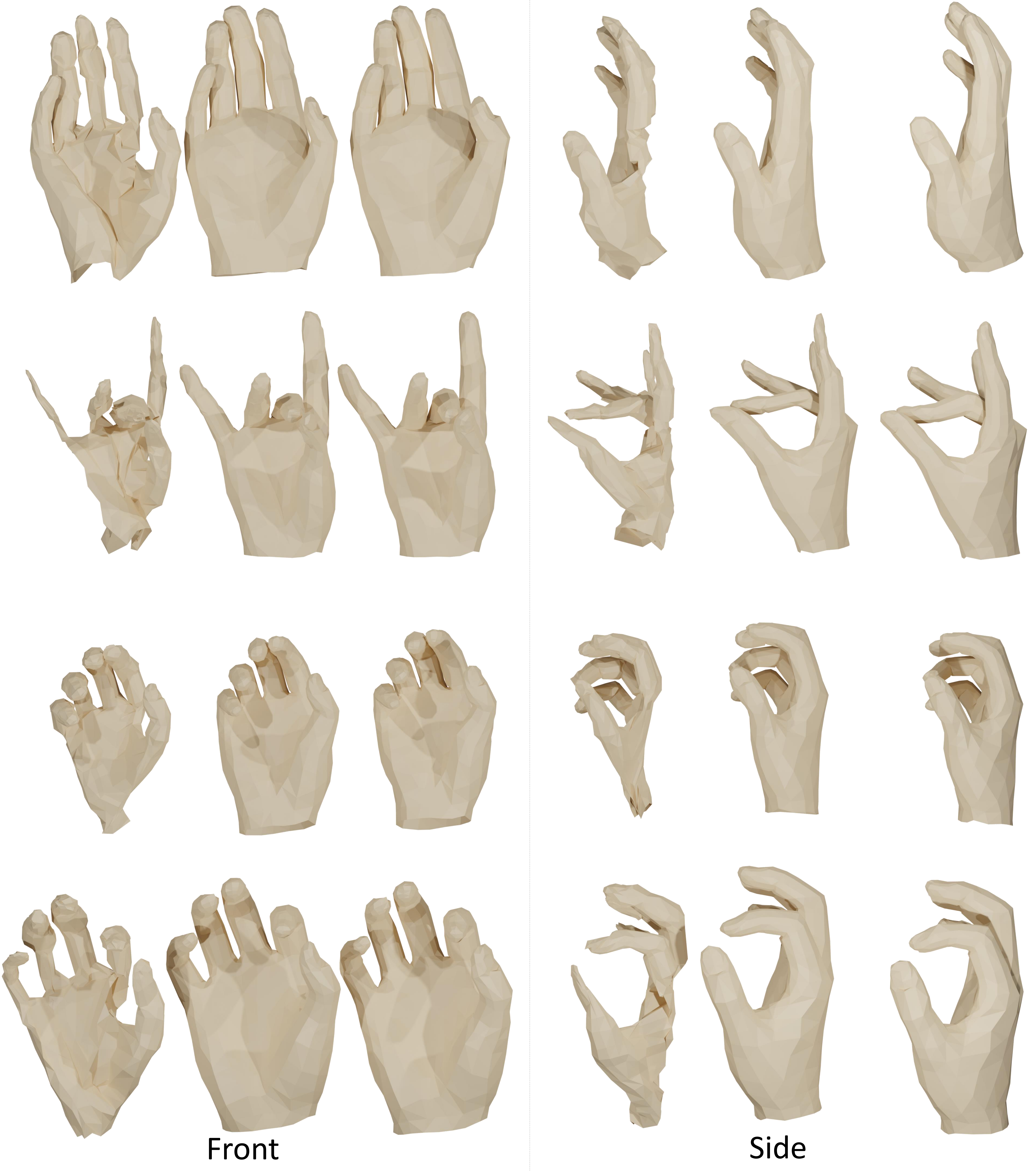}
    \caption{The qualitative display of the ablation of GSD module. All examples come from the FreiHand test set. The left half presents the frontal view, while the right half displays the side view. In each half, moving from left to right, the initial column showcases outcomes without GSD, the middle column exhibits results with GSD, and the final column portrays the ground-truth meshes.}
    \label{fig:gsd_ablation}
\end{figure}

\noindent{\bf The design of the Skeleton2Mesh model.}
We conduct a series of ablation studies to illustrate the rationale behind the design of the Skeleton2Mesh model.
And all following ablation experiments are conducted in FreiHand~\cite{zimmermann2019freihand} dataset.
To explore the influence of the depth of MLP $f_{{\beta}_i}(\cdot)$, we vary the depth from 2 to 5 and figure out that our Skeleton2Mesh model  exhibits a boundary effect in terms of network depth. 
As reported in~\cref{tab:ablation_basic_mlphand}, as the number of layers decreases, the model's performance marginally degrades. However, when the number of layers reduces to two, its performance collapses abruptly.
This experiments reveals that employing a three-layer fully connected architecture is a balanced choice between precision and computational cost.
To explore the influence of the Position Encoding, Global Spatial Descriptor, and the per-bone joint learning strategy.
we compare the results of \textit{w} and \textit{w/o} these components in the 3-layers MLP setting.
As represented in~\cref{tab:ablation_basic_mlphand}, we  find: the Position Encoding can improve reconstruction accuracy; the absence of the GSD module significantly reduces accuracy metrics and leads to the generation of heavy geometric artifacts in the palm as displayed in \cref{fig:gsd_ablation}; and the joint learning strategy contains good accuracy performance while has less parameters. 
The detailed explaination of how GSD works and the ablation study of two-stage training strategy are placed in \textit{supplemental material}, due to the page limitation.



\section{Conclusion}
In this paper, we introduce MLPHand, a novel approach to real-time multi-view hand mesh reconstruction. The primary goal of this approach is to enhance the inference speed of the network without compromising the reconstruction accuracy. To achieve this, we meticulously design two pivotal components within MLPHand: the Skeleton2Mesh model and the multi-view geometry feature fusion prediction module, which both consisting of only MLP layers. We have conducted extensive experiments on three widely recognized multi-view hand reconstruction datasets to validate our proposed method. The results from these empirical assessments underscore the effectiveness and efficiency of MLPHand, affirming its potential as a practical solution for real-time multi-view hand reconstruction tasks.

{
    \small
    \bibliographystyle{ieeenat_fullname}
    \bibliography{main}

\begin{thebibliography}{10}

\bibitem{yang2023poem}
Lixin Yang, Jian Xu, Licheng Zhong, Xinyu Zhan, Zhicheng Wang, Kejian Wu, and Cewu Lu.
\newblock Poem: Reconstructing hand in a point embedded multi-view stereo.
\newblock In {\em Proceedings of the IEEE/CVF Conference on Computer Vision and Pattern Recognition}, pages 21108--21117, 2023.

\bibitem{zhang2023adding}
Lvmin Zhang, Anyi Rao, and Maneesh Agrawala.
\newblock Adding conditional control to text-to-image diffusion models.
\newblock In {\em Proceedings of the IEEE/CVF International Conference on Computer Vision}, pages 3836--3847, 2023.

\bibitem{karunratanakul2021skeleton}
Korrawe Karunratanakul, Adrian Spurr, Zicong Fan, Otmar Hilliges, and Siyu Tang.
\newblock A skeleton-driven neural occupancy representation for articulated hands.
\newblock In {\em 2021 International Conference on 3D Vision (3DV)}, pages 11--21. IEEE, 2021.

\bibitem{deng2020nasa}
Boyang Deng, John~P Lewis, Timothy Jeruzalski, Gerard Pons-Moll, Geoffrey Hinton, Mohammad Norouzi, and Andrea Tagliasacchi.
\newblock Nasa neural articulated shape approximation.
\newblock In {\em Computer Vision--ECCV 2020: 16th European Conference, Glasgow, UK, August 23--28, 2020, Proceedings, Part VII 16}, pages 612--628. Springer, 2020.

\bibitem{romero2022embodied}
Javier Romero, Dimitrios Tzionas, and Michael~J Black.
\newblock Embodied hands: Modeling and capturing hands and bodies together.
\newblock {\em arXiv preprint arXiv:2201.02610}, 2022.

\bibitem{boukhayma20193d}
Adnane Boukhayma, Rodrigo~de Bem, and Philip~HS Torr.
\newblock 3d hand shape and pose from images in the wild.
\newblock In {\em Proceedings of the IEEE/CVF Conference on Computer Vision and Pattern Recognition}, pages 10843--10852, 2019.

\bibitem{hasson2019learning}
Yana Hasson, Gul Varol, Dimitrios Tzionas, Igor Kalevatykh, Michael~J Black, Ivan Laptev, and Cordelia Schmid.
\newblock Learning joint reconstruction of hands and manipulated objects.
\newblock In {\em Proceedings of the IEEE/CVF conference on computer vision and pattern recognition}, pages 11807--11816, 2019.

\bibitem{kong2022identity}
Deying Kong, Linguang Zhang, Liangjian Chen, Haoyu Ma, Xiangyi Yan, Shanlin Sun, Xingwei Liu, Kun Han, and Xiaohui Xie.
\newblock Identity-aware hand mesh estimation and personalization from rgb images.
\newblock In {\em European Conference on Computer Vision}, pages 536--553. Springer, 2022.

\bibitem{zimmermann2019freihand}
Christian Zimmermann, Duygu Ceylan, Jimei Yang, Bryan Russell, Max Argus, and Thomas Brox.
\newblock Freihand: A dataset for markerless capture of hand pose and shape from single rgb images.
\newblock In {\em Proceedings of the IEEE/CVF International Conference on Computer Vision}, pages 813--822, 2019.

\bibitem{zhou2020monocular}
Yuxiao Zhou, Marc Habermann, Weipeng Xu, Ikhsanul Habibie, Christian Theobalt, and Feng Xu.
\newblock Monocular real-time hand shape and motion capture using multi-modal data.
\newblock In {\em Proceedings of the IEEE/CVF Conference on Computer Vision and Pattern Recognition}, pages 5346--5355, 2020.

\bibitem{chen2021camera}
Xingyu Chen, Yufeng Liu, Chongyang Ma, Jianlong Chang, Huayan Wang, Tian Chen, Xiaoyan Guo, Pengfei Wan, and Wen Zheng.
\newblock Camera-space hand mesh recovery via semantic aggregation and adaptive 2d-1d registration.
\newblock In {\em Proceedings of the IEEE/CVF Conference on Computer Vision and Pattern Recognition}, pages 13274--13283, 2021.

\bibitem{tse2022collaborative}
Tze Ho~Elden Tse, Kwang~In Kim, Ales Leonardis, and Hyung~Jin Chang.
\newblock Collaborative learning for hand and object reconstruction with attention-guided graph convolution.
\newblock In {\em Proceedings of the IEEE/CVF Conference on Computer Vision and Pattern Recognition}, pages 1664--1674, 2022.

\bibitem{ge20193d}
Liuhao Ge, Zhou Ren, Yuncheng Li, Zehao Xue, Yingying Wang, Jianfei Cai, and Junsong Yuan.
\newblock 3d hand shape and pose estimation from a single rgb image.
\newblock In {\em Proceedings of the IEEE/CVF Conference on Computer Vision and Pattern Recognition}, pages 10833--10842, 2019.

\bibitem{kolotouros2019convolutional}
Nikos Kolotouros, Georgios Pavlakos, and Kostas Daniilidis.
\newblock Convolutional mesh regression for single-image human shape reconstruction.
\newblock In {\em Proceedings of the IEEE/CVF Conference on Computer Vision and Pattern Recognition}, pages 4501--4510, 2019.

\bibitem{kulon2020weakly}
Dominik Kulon, Riza~Alp Guler, Iasonas Kokkinos, Michael~M Bronstein, and Stefanos Zafeiriou.
\newblock Weakly-supervised mesh-convolutional hand reconstruction in the wild.
\newblock In {\em Proceedings of the IEEE/CVF conference on computer vision and pattern recognition}, pages 4990--5000, 2020.

\bibitem{chen2022mobrecon}
Xingyu Chen, Yufeng Liu, Yajiao Dong, Xiong Zhang, Chongyang Ma, Yanmin Xiong, Yuan Zhang, and Xiaoyan Guo.
\newblock Mobrecon: Mobile-friendly hand mesh reconstruction from monocular image.
\newblock In {\em Proceedings of the IEEE/CVF Conference on Computer Vision and Pattern Recognition}, pages 20544--20554, 2022.

\bibitem{wang2020rgb2hands}
Jiayi Wang, Franziska Mueller, Florian Bernard, Suzanne Sorli, Oleksandr Sotnychenko, Neng Qian, Miguel~A Otaduy, Dan Casas, and Christian Theobalt.
\newblock Rgb2hands: real-time tracking of 3d hand interactions from monocular rgb video.
\newblock {\em ACM Transactions on Graphics (ToG)}, 39(6):1--16, 2020.

\bibitem{lin2021mesh}
Kevin Lin, Lijuan Wang, and Zicheng Liu.
\newblock Mesh graphormer.
\newblock In {\em Proceedings of the IEEE/CVF international conference on computer vision}, pages 12939--12948, 2021.

\bibitem{chen2021i2uv}
Ping Chen, Yujin Chen, Dong Yang, Fangyin Wu, Qin Li, Qingpei Xia, and Yong Tan.
\newblock I2uv-handnet: Image-to-uv prediction network for accurate and high-fidelity 3d hand mesh modeling.
\newblock In {\em Proceedings of the IEEE/CVF International Conference on Computer Vision}, pages 12929--12938, 2021.

\bibitem{lin2021end}
Kevin Lin, Lijuan Wang, and Zicheng Liu.
\newblock End-to-end human pose and mesh reconstruction with transformers.
\newblock In {\em Proceedings of the IEEE/CVF conference on computer vision and pattern recognition}, pages 1954--1963, 2021.

\bibitem{chen2022alignsdf}
Zerui Chen, Yana Hasson, Cordelia Schmid, and Ivan Laptev.
\newblock Alignsdf: Pose-aligned signed distance fields for hand-object reconstruction.
\newblock In {\em European Conference on Computer Vision}, pages 231--248. Springer, 2022.

\bibitem{corona2022lisa}
Enric Corona, Tomas Hodan, Minh Vo, Francesc Moreno-Noguer, Chris Sweeney, Richard Newcombe, and Lingni Ma.
\newblock Lisa: Learning implicit shape and appearance of hands.
\newblock In {\em Proceedings of the IEEE/CVF Conference on Computer Vision and Pattern Recognition}, pages 20533--20543, 2022.

\bibitem{chen2023hand}
Xingyu Chen, Baoyuan Wang, and Heung-Yeung Shum.
\newblock Hand avatar: Free-pose hand animation and rendering from monocular video.
\newblock In {\em Proceedings of the IEEE/CVF Conference on Computer Vision and Pattern Recognition}, pages 8683--8693, 2023.

\bibitem{mildenhall2021nerf}
Ben Mildenhall, Pratul~P Srinivasan, Matthew Tancik, Jonathan~T Barron, Ravi Ramamoorthi, and Ren Ng.
\newblock Nerf: Representing scenes as neural radiance fields for view synthesis.
\newblock {\em Communications of the ACM}, 65(1):99--106, 2021.

\bibitem{mescheder2019occupancy}
Lars Mescheder, Michael Oechsle, Michael Niemeyer, Sebastian Nowozin, and Andreas Geiger.
\newblock Occupancy networks: Learning 3d reconstruction in function space.
\newblock In {\em Proceedings of the IEEE/CVF conference on computer vision and pattern recognition}, pages 4460--4470, 2019.

\bibitem{sun2018integral}
Xiao Sun, Bin Xiao, Fangyin Wei, Shuang Liang, and Yichen Wei.
\newblock Integral human pose regression.
\newblock In {\em Proceedings of the European conference on computer vision (ECCV)}, pages 529--545, 2018.

\bibitem{hartley2003multiple}
Richard Hartley and Andrew Zisserman.
\newblock {\em Multiple view geometry in computer vision}.
\newblock Cambridge university press, 2003.

\bibitem{mihajlovic2022coap}
Marko Mihajlovic, Shunsuke Saito, Aayush Bansal, Michael Zollhoefer, and Siyu Tang.
\newblock Coap: Compositional articulated occupancy of people.
\newblock In {\em Proceedings of the IEEE/CVF Conference on Computer Vision and Pattern Recognition}, pages 13201--13210, 2022.

\bibitem{kingma2014adam}
Diederik~P Kingma and Jimmy Ba.
\newblock Adam: A method for stochastic optimization.
\newblock {\em arXiv preprint arXiv:1412.6980}, 2014.

\bibitem{russakovsky2015imagenet}
Olga Russakovsky, Jia Deng, Hao Su, Jonathan Krause, Sanjeev Satheesh, Sean Ma, Zhiheng Huang, Andrej Karpathy, Aditya Khosla, Michael Bernstein, et~al.
\newblock Imagenet large scale visual recognition challenge.
\newblock {\em International journal of computer vision}, 115:211--252, 2015.

\bibitem{he2016deep}
Kaiming He, Xiangyu Zhang, Shaoqing Ren, and Jian Sun.
\newblock Deep residual learning for image recognition.
\newblock In {\em Proceedings of the IEEE conference on computer vision and pattern recognition}, pages 770--778, 2016.

\bibitem{sandler2018mobilenetv2}
Mark Sandler, Andrew Howard, Menglong Zhu, Andrey Zhmoginov, and Liang-Chieh Chen.
\newblock Mobilenetv2: Inverted residuals and linear bottlenecks.
\newblock In {\em Proceedings of the IEEE conference on computer vision and pattern recognition}, pages 4510--4520, 2018.

\bibitem{chao2021dexycb}
Yu-Wei Chao, Wei Yang, Yu~Xiang, Pavlo Molchanov, Ankur Handa, Jonathan Tremblay, Yashraj~S Narang, Karl Van~Wyk, Umar Iqbal, Stan Birchfield, et~al.
\newblock Dexycb: A benchmark for capturing hand grasping of objects.
\newblock In {\em Proceedings of the IEEE/CVF Conference on Computer Vision and Pattern Recognition}, pages 9044--9053, 2021.

\bibitem{hampali2021ho}
Shreyas Hampali, Sayan~Deb Sarkar, and Vincent Lepetit.
\newblock Ho-3d\_v3: Improving the accuracy of hand-object annotations of the ho-3d dataset.
\newblock {\em arXiv preprint arXiv:2107.00887}, 2021.

\bibitem{yang2022oakink}
Lixin Yang, Kailin Li, Xinyu Zhan, Fei Wu, Anran Xu, Liu Liu, and Cewu Lu.
\newblock Oakink: A large-scale knowledge repository for understanding hand-object interaction.
\newblock In {\em Proceedings of the IEEE/CVF Conference on Computer Vision and Pattern Recognition}, pages 20953--20962, 2022.

\bibitem{gower1975generalized}
John~C Gower.
\newblock Generalized procrustes analysis.
\newblock {\em Psychometrika}, 40:33--51, 1975.

\bibitem{zhang2021direct}
Jianfeng Zhang, Yujun Cai, Shuicheng Yan, Jiashi Feng, et~al.
\newblock Direct multi-view multi-person 3d pose estimation.
\newblock {\em Advances in Neural Information Processing Systems}, 34:13153--13164, 2021.

\bibitem{Li_2021_WACV}
Zhongguo Li, Magnus Oskarsson, and Anders Heyden.
\newblock 3d human pose and shape estimation through collaborative learning and multi-view model-fitting.
\newblock In {\em Proceedings of the IEEE/CVF Winter Conference on Applications of Computer Vision (WACV)}, pages 1888--1897, January 2021.

\bibitem{buckingham2021hand}
Gavin Buckingham.
\newblock Hand tracking for immersive virtual reality: opportunities and challenges.
\newblock {\em Frontiers in Virtual Reality}, 2:728461, 2021.

\bibitem{kumar2015mujoco}
Vikash Kumar and Emanuel Todorov.
\newblock Mujoco haptix: A virtual reality system for hand manipulation.
\newblock In {\em 2015 IEEE-RAS 15th International Conference on Humanoid Robots (Humanoids)}, pages 657--663. IEEE, 2015.

\bibitem{reifinger2007static}
Stefan Reifinger, Frank Wallhoff, Markus Ablassmeier, Tony Poitschke, and Gerhard Rigoll.
\newblock Static and dynamic hand-gesture recognition for augmented reality applications.
\newblock In {\em Human-Computer Interaction. HCI Intelligent Multimodal Interaction Environments: 12th International Conference, HCI International 2007, Beijing, China, July 22-27, 2007, Proceedings, Part III 12}, pages 728--737. Springer, 2007.

\bibitem{radkowski2012interactive}
Rafael Radkowski and Christian Stritzke.
\newblock Interactive hand gesture-based assembly for augmented reality applications.
\newblock In {\em Proceedings of the 2012 International Conference on Advances in Computer-Human Interactions}, pages 303--308. Citeseer, 2012.

\bibitem{fu2022geo}
Qiancheng Fu, Qingshan Xu, Yew~Soon Ong, and Wenbing Tao.
\newblock Geo-neus: Geometry-consistent neural implicit surfaces learning for multi-view reconstruction.
\newblock {\em Advances in Neural Information Processing Systems}, 35:3403--3416, 2022.

\bibitem{yu2022monosdf}
Zehao Yu, Songyou Peng, Michael Niemeyer, Torsten Sattler, and Andreas Geiger.
\newblock Monosdf: Exploring monocular geometric cues for neural implicit surface reconstruction.
\newblock {\em Advances in neural information processing systems}, 35:25018--25032, 2022.

\bibitem{park2019deepsdf}
Jeong~Joon Park, Peter Florence, Julian Straub, Richard Newcombe, and Steven Lovegrove.
\newblock Deepsdf: Learning continuous signed distance functions for shape representation.
\newblock In {\em Proceedings of the IEEE/CVF conference on computer vision and pattern recognition}, pages 165--174, 2019.

\bibitem{chou2023diffusion}
Gene Chou, Yuval Bahat, and Felix Heide.
\newblock Diffusion-sdf: Conditional generative modeling of signed distance functions.
\newblock In {\em Proceedings of the IEEE/CVF International Conference on Computer Vision}, pages 2262--2272, 2023.

\bibitem{wang2022clip}
Can Wang, Menglei Chai, Mingming He, Dongdong Chen, and Jing Liao.
\newblock Clip-nerf: Text-and-image driven manipulation of neural radiance fields.
\newblock In {\em Proceedings of the IEEE/CVF Conference on Computer Vision and Pattern Recognition}, pages 3835--3844, 2022.

\bibitem{hong2022headnerf}
Yang Hong, Bo~Peng, Haiyao Xiao, Ligang Liu, and Juyong Zhang.
\newblock Headnerf: A real-time nerf-based parametric head model.
\newblock In {\em Proceedings of the IEEE/CVF Conference on Computer Vision and Pattern Recognition}, pages 20374--20384, 2022.

\bibitem{pumarola2021d}
Albert Pumarola, Enric Corona, Gerard Pons-Moll, and Francesc Moreno-Noguer.
\newblock D-nerf: Neural radiance fields for dynamic scenes.
\newblock In {\em Proceedings of the IEEE/CVF Conference on Computer Vision and Pattern Recognition}, pages 10318--10327, 2021.

\bibitem{wu2020multi}
Minye Wu, Yuehao Wang, Qiang Hu, and Jingyi Yu.
\newblock Multi-view neural human rendering.
\newblock In {\em Proceedings of the IEEE/CVF Conference on Computer Vision and Pattern Recognition}, pages 1682--1691, 2020.

\bibitem{gordon2022flex}
Brian Gordon, Sigal Raab, Guy Azov, Raja Giryes, and Daniel Cohen-Or.
\newblock Flex: Extrinsic parameters-free multi-view 3d human motion reconstruction.
\newblock In {\em European Conference on Computer Vision}, pages 176--196. Springer, 2022.

\bibitem{dong2021shape}
Zijian Dong, Jie Song, Xu~Chen, Chen Guo, and Otmar Hilliges.
\newblock Shape-aware multi-person pose estimation from multi-view images.
\newblock In {\em Proceedings of the IEEE/CVF International Conference on Computer Vision}, pages 11158--11168, 2021.

\bibitem{Tse_2023_ICCV}
Tze Ho~Elden Tse, Franziska Mueller, Zhengyang Shen, Danhang Tang, Thabo Beeler, Mingsong Dou, Yinda Zhang, Sasa Petrovic, Hyung~Jin Chang, Jonathan Taylor, and Bardia Doosti.
\newblock Spectral graphormer: Spectral graph-based transformer for egocentric two-hand reconstruction using multi-view color images.
\newblock In {\em Proceedings of the IEEE/CVF International Conference on Computer Vision (ICCV)}, pages 14666--14677, October 2023.

\bibitem{saito2019pifu}
Shunsuke Saito, Zeng Huang, Ryota Natsume, Shigeo Morishima, Angjoo Kanazawa, and Hao Li.
\newblock Pifu: Pixel-aligned implicit function for high-resolution clothed human digitization.
\newblock In {\em Proceedings of the IEEE/CVF international conference on computer vision}, pages 2304--2314, 2019.

\bibitem{xu2023h2onet}
Hao Xu, Tianyu Wang, Xiao Tang, and Chi-Wing Fu.
\newblock H2onet: Hand-occlusion-and-orientation-aware network for real-time 3d hand mesh reconstruction.
\newblock In {\em Proceedings of the IEEE/CVF Conference on Computer Vision and Pattern Recognition}, pages 17048--17058, 2023.

\bibitem{wei2016convolutional}
Shih-En Wei, Varun Ramakrishna, Takeo Kanade, and Yaser Sheikh.
\newblock Convolutional pose machines.
\newblock In {\em Proceedings of the IEEE conference on Computer Vision and Pattern Recognition}, pages 4724--4732, 2016.

\end{thebibliography}
}
{
\clearpage
\setcounter{page}{1}

\section{Loss functions}
In the training of the Light-weight Skeleton2Mesh model. 
We use L2 norm for the 3D mesh loss:
\begin{equation}
    \mathcal{L}_{mesh}^1=||{\bf \hat{V}}-f_{\beta}({\bf \hat{X}})||_2
    \label{eq:loss_}
\end{equation}
where ${\bf \hat{V}}$ is the ground-truth mesh, ${\bf \hat{X}}$ is the ground-truth skeleton, and $\beta$ is the parameter of the Skeleton2Mesh model.
In the training of the whole pipeline, we use L2 norm for skeleton-level loss and L1 norm for vertex-level loss. 
Specifically, we have:
\begin{equation}
    \begin{array}{c}
    \mathcal{L}_{heatmap}^{2D}=\sum_{n=1}^N ||{\bf \hat{H}}_n-{\bf {H}}_n ||_2\\
    ~\\
    \mathcal{L}_{skeleton}^{2D}=\sum_{n=1}^N ||{\bf \hat{X}}_n^{2D}-{\bf {X}}_n^{2D}||_2 \\~\\
    \mathcal{L}_{skeleton}^{3D}=||{\bf \hat{X}}-{\bf {X}}||_2 \\~\\
    \mathcal{L}_{vertex}^{2D}=\sum_{n=1}^N ||{\bf \hat{V}}_n^{2D}-{\bf {V}}_n^{2D}||_1  \\~\\
    \mathcal{L}_{vertex}^{3D}=||{\bf \hat{V}}-{\bf {V}}||_1 \\~\\
    \end{array}
\end{equation}
where, $N$ is the number of camera views, $\hat{(\cdot)}$ indicates the ground-truth counterpart, the ${\bf H}_n$ means the heatmap of ${\bf X}_n^{2D}$ , and  ${(\cdot)}^{2D}$ means the projection from 3D to 2D.
We use coefficients to balance these above loss items.
Formally, the objective function in the second training stage is: 
\begin{equation}
\begin{aligned}
  \mathcal{L}=& \lambda_1 \cdot \mathcal{L}_{heatmap}^{2D} + \lambda_2 \cdot  \mathcal{L}_{skeleton}^{2D} \\
& \lambda_3 \cdot  \mathcal{L}_{vertex}^{2D}+  \lambda_4 \cdot \mathcal{L}_{skeleton}^{3D} +\lambda_5 \cdot \mathcal{L}_{vertex}^{3D},  
\end{aligned}
\end{equation}
where empirically $\lambda_1=10$, $\lambda_2=\lambda_4=\lambda_5=1$, and $\lambda_3=0.1$.

\section{More Implementation Details}
In our Skelton2Mesh model and the Global Spatial Descriptor (GSD), we set 256 neurons in hidden layers, and each fully-connected (FC) layer includes the \textit{LeakyReLu} activation, except for the last one.
The GSD consists of two FC layers with \textit{LeakyReLu}, and outputs an 100 dimensional vector, and takes all non-root keypoints' coordinate as input, ${\rm GSD}(\cdot): \mathbb{R}^{21 \times 3} \rightarrow \mathbb{R}^{100}$.
In the \textbf{convex decomposition} of hands, the vertex numbers of 20 local meshes are 45, 61, 43, 45, 92, 34, 41, 62, 44, 44, 58, 42, 40, 60, 41, 35, 64, 28, 50, 62, respectively (Sum is 991). 
For achieving decomposition, we use a decomposition matrix, ${\bf M} \in \mathbb{R}^{991\times 778}$, to left-multiply the vertex matrix, ${\bf V} \in \mathbb{R}^{778\times 3}$.
And we use the left inverse of  ${\bf M}$ to recover ${\bf V}$ by left-multiplication.
Besides, the output layers have 100 neurons, and we only compute the loss in part of neurons, where the number of these neurons is same with the vertex number of the corresponding local mesh. 

\section{The 6D Pose of Each Bone}
\label{6dpose}
In this section, we prove that the 6D pose of each bone is determined by the two endpoints of  per-bone, and  we use a articulated object to explain this as the articulation nature of  hand skeletons. As depicted in \cref{fig:supple_gsd},  in case of the green part, we can obtain a local coordinate system by  three points ${\bf A}$, ${\bf B}$, ${\bf O}$, and the 6D pose of the green part can be obtained from the relationship between this  local coordinate system and the world coordinate system.
Specifically, in local coordinate system, we set the child node (the point ${\bf B}$) as the origin, the direction from ${\bf B}$ to ${\bf A}$ as the x-axis, the direction from ${\bf B}$ to ${\bf O}$ as the y-axis, and the cross product of  x-axis and  y-axis as the z-axis. 

However, in the yellow part (as the palm part), its 6D pose can be obtained following above way, as the overlap between the child node ${\bf C}$ and the wrist point ${\bf O}$. 

\begin{figure*}[!h]
    \centering
    \includegraphics[width=1.0\columnwidth]{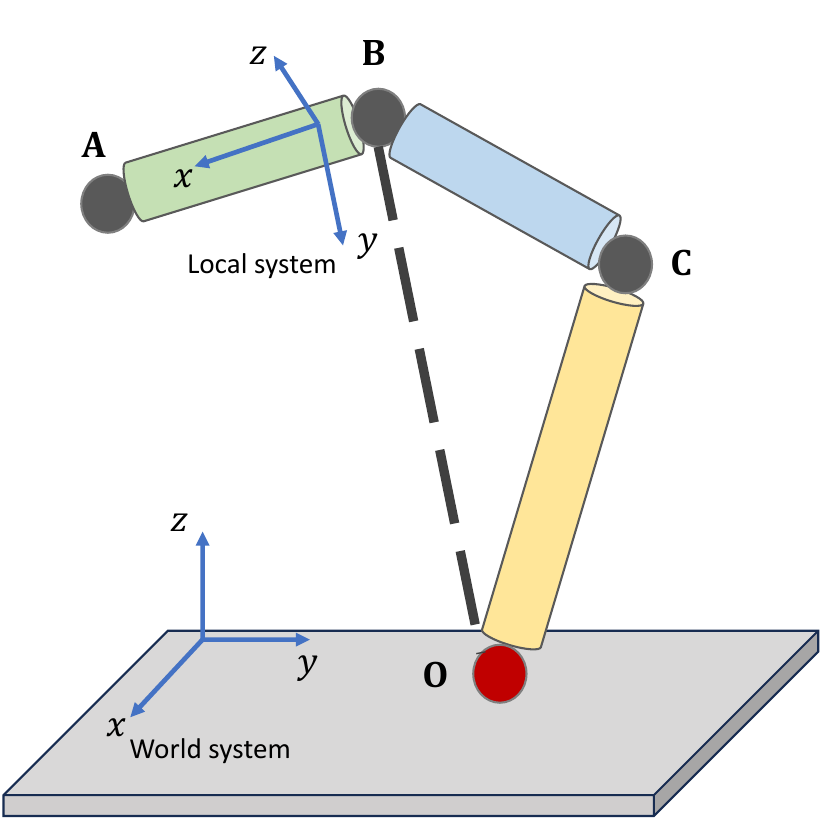}
    \caption{The illustration of the computation of the 6D pose of per-bone.}
    \label{fig:supple_gsd}
\end{figure*}

\section{Explanation of GSD}
We introduce the GSD aiming to offer extra spatial information to alleviate the inadequacy of spatial information discussed in \cref{6dpose}. 
Specifically, ${\rm GSD}(\cdot): \mathbb{R}^{21 \times 3} \rightarrow \mathbb{R}^{100}$ is a two-layer MLP network.
It takes the skeletons as input and outputs an supplemental spatial vector.
In the main text, we qualitatively demonstrate the efficacy of the GSD in Skeleton2Mesh model, and we quantitatively demonstrate it in \cref{tab:supple_gsd} with more details.
As reported in this table, we display per-bone reconstruction metrics in hand palm. 
It turns out that  this straightforward strategy can apparently alleviate the degeneration of the root-relative setting: there are remarkable accuracy improvements on the palm region (the 1 to 5 part), which demonstrates the effect of the GSD.
\begin{table}[h]
    \centering
    \begin{tabular}{c|c|c|c|c|c}
    \toprule
      GSD  & 1-st &2-nd &3-rd &4-th &5-th   \\
      \midrule
      $\checkmark$   & \textbf{1.44} & \textbf{1.35} & \textbf{1.22} & \textbf{1.15} & \textbf{1.23}\\
      $\times$ &18.05&13.99&13.15&11.69&14.00\\
      \bottomrule
    \end{tabular}
    \caption{The quantitative experiment of the GSD. We report the mean per vertex position error (MPVPE) in this table, where $\checkmark$ and $\times$ indicate the experiment \textit{w} and \textit{w/o} the GSD respectively.  }
    \label{tab:supple_gsd}
\end{table}

\section{Ablation study of training strategy}
As we claim in main text, the two-stage training allows better accuracy. 
We demonstrate this under the experiment reported in \cref{tab:supple_train_strategy}.
The $\checkmark$ in "Two Stage Training" column indicates training MLPHand under in two stage, and $\times$ indicates training MLPHand just in an end-to-end way. 
This experiment is conducted in Oaklnk-MV dataset, and the CNN backbone is ResNet-34.

\begin{table}[h]
    \centering
    \begin{tabular}{c|c|c}
    \toprule
      Two Stage Training  & MPJPE & MPVPE   \\
      \midrule
      $\checkmark$   & \textbf{6.17} & \textbf{6.00}\\
      $\times$ &6.53&6.64\\
      \bottomrule
    \end{tabular}
    \caption{The quantitative experiment of the training strategy. We report the mean per vertex/joint position error (MPVPE/MPJPE)  in this table, where $\checkmark$ and $\times$ indicate the experiment "do" and "do not" using two-stage training strategy.  }
    \label{tab:supple_train_strategy}
\end{table}

\section{Additional Experiments}
To make our comparison more comprehensive, in \cref{tab:addition}, we additionally compare our MLPHand with recent efficient SOTA of monocular hand reconstruction methods\cite{chen2022mobrecon,xu2023h2onet} to demonstrate the accuracy advantage of multi-view hand reconstruction method.
Besides, considering that egocentric scenario is important, hence we report the result of 2 close view setting in \cref{tab:addition}. Close 2-views setting is close to the egocentric scenario and more challenging. The result turns out that the accuracy of MLPHand hasn't collapse in more challenging 2-view setting, indicating its potential on egocentric scenario.
\begin{table}[h]
    \centering\resizebox{0.50\textwidth}{!}{
    \begin{tabular}{c|c|c|c|c|c|c}
    \toprule
    Method& MPJPE & MPVPE & Multi-ADDs & Params& Backbone& FPS\\
    \hline
     MobRecon\cite{chen2022mobrecon}&14.4&13.3&0.5G&  8.2M& DenseStack & 89\\
     H2ONet\cite{xu2023h2onet}    &14.2&13.2&0.7G&25.9M& DenseStack &73\\     
     \hline
     Ours    &6.8&6.9&6.1G&12.5M& MobileNetV2& 63\\
     Ours (2-views)    &10.6&10.2&13.2G&25.8M& ResNet34 &70\\
     
    \midrule
     
    \end{tabular}}
    \caption{Additional comparison on Dex-YCB dataset. All experiments have same data split, excluding the 2-view experiment. This experiment will be added to our supplemental material. Apparently, MVHMR methods outperforms MHR methods in terms of accuracy,  because of the usage of multi-view images. Note the DenseStack and MobileNetV2 are more friendly with CPU
    device.}
    \label{tab:addition}
\end{table}

\section{Qualitative results}
We demonstrate more qualitative results of MLPHand on the three datasets in \cref{fig:quali_dexycb,fig:quali_ho3d,fig:quali_oaklnk}. From top to bottom we plot the results of different CNN backbone on DexYCB-MV, HO3D-MV and OakInk-MV datasets. 
For each multi-view frame, we draw its result from 5 different views. One of the views is in normal size and the other four views are half size. 
\begin{figure*}[h]
    \centering
    \includegraphics[width=1.0\textwidth]{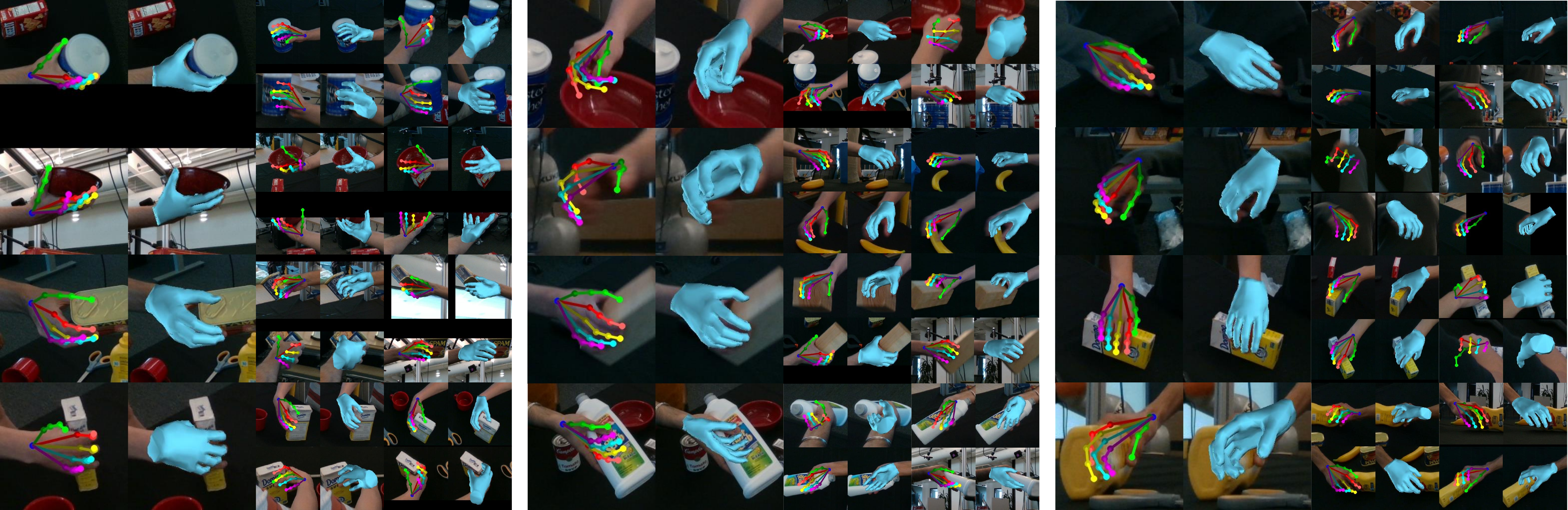}
    \caption{ The qualitative results on DexYCB-MV testing set. The CNN backbone in this experiment is ResNet-34\cite{he2016deep}}
    \label{fig:quali_dexycb}
\end{figure*}

\begin{figure*}[h]
    \centering
    \includegraphics[width=1.0\textwidth]{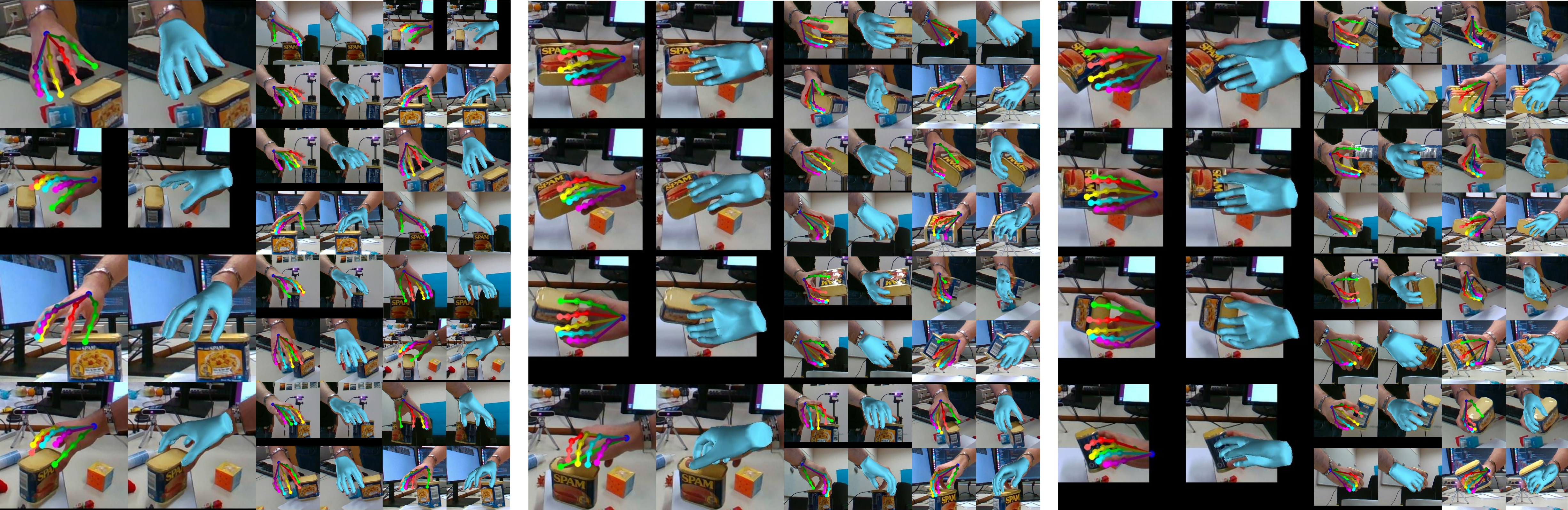}
    \caption{ The qualitative results on HO3DV3-MV testing set. The CNN backbone in this experiment is ResNet-18~\cite{he2016deep}}
    \label{fig:quali_ho3d}
\end{figure*}

\begin{figure*}[h]
    \centering
    \includegraphics[width=1.0\textwidth]{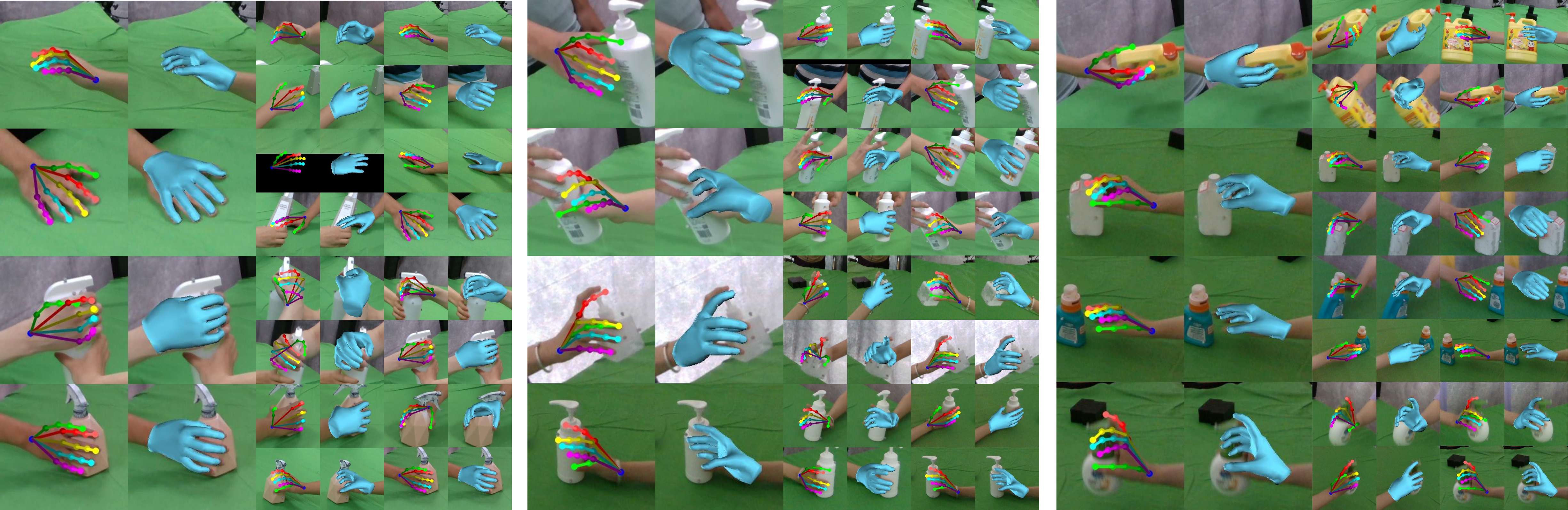}
    \caption{ The qualitative results on Oaklnk-MV testing set. The CNN backbone in this experiment is MobileNetv2~\cite{sun2018integral}}
    \label{fig:quali_oaklnk}
\end{figure*}
\clearpage

}

\end{document}